\pdfoutput=1

\documentclass[11pt]{article}
\usepackage[dvipsnames]{xcolor}

\usepackage[preprint]{acl}

\usepackage{times}
\usepackage{latexsym}

\usepackage[T1]{fontenc}

\usepackage[utf8]{inputenc}

\usepackage{microtype}

\usepackage{inconsolata}

\usepackage{graphicx}
\usepackage{tabularx}
\usepackage{multirow}
\usepackage{amsfonts}
\usepackage{amssymb}
\usepackage{xspace}
\usepackage{lipsum}
\usepackage{adjustbox}
\usepackage{booktabs}
\usepackage{makecell}
\usepackage{url}
\usepackage{enumitem}
\setlist[enumerate]{topsep=0pt, partopsep=0pt, parsep=0pt, itemsep=0pt}
\setlist[itemize]{topsep=0pt, partopsep=0pt, parsep=0pt, itemsep=0pt}
\usepackage{listings}
\usepackage{xcolor}
\usepackage{inconsolata}
\usepackage[most]{tcolorbox}
\usepackage{caption}
\usepackage{float}
\usepackage{pgfplots}
\pgfplotsset{compat=1.17}
\usepackage{pgfplotstable}
\usepackage{amsthm}
\newtheorem{definition}{Definition}
\newtheorem{theorem}{Theorem}

\newcommand{\cope}{$\texttt{CoPE}$\xspace}
\newcommand{\df}{$\mathcal{DF}$\xspace}
\newcommand{\pf}{$\mathcal{PF}$\xspace}

\newcommand{\bw}{$\texttt{BlocksWorld-100}$\xspace}
\newcommand{\mbw}{$\texttt{MysteryBlocksWorld-100}$\xspace}
\newcommand{\bwxl}{$\texttt{BlocksWorld-XL--100}$\xspace}

\newcommand{\coin}{$\texttt{CoinCollector-100}$\xspace}

\newcommand{\drexel}{%
  \hspace{1pt}
  \begingroup\normalfont
  \includegraphics[height=1.3\fontcharht\font`\B]{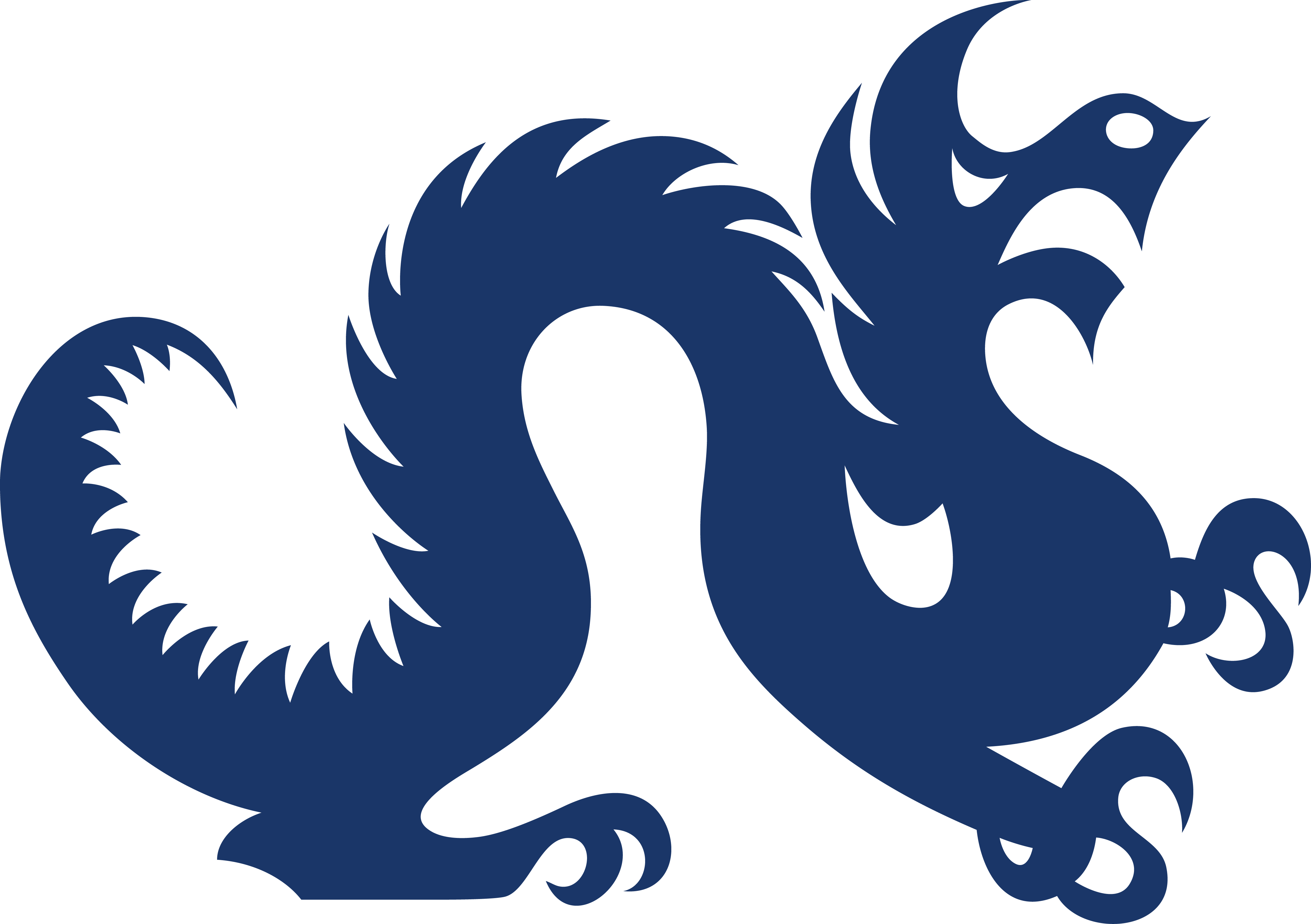}%
  \endgroup
  \hspace{1pt}
}
\newcommand{\brown}{%
  \hspace{1pt}
  \begingroup\normalfont
  \includegraphics[height=1.3\fontcharht\font`\B]{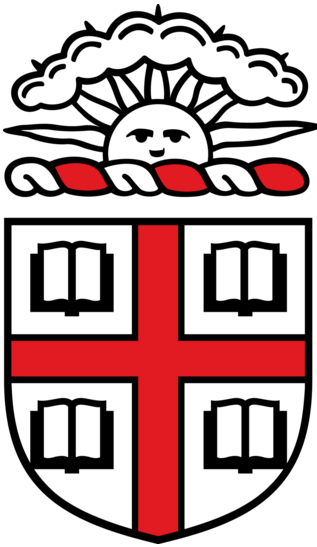}%
  \endgroup
  \hspace{1pt}
}

\lstdefinestyle{mystyle}{
    basicstyle=\ttfamily\small,
    breaklines=true,
    frame=single,
    backgroundcolor=\color{gray!5},
    keywordstyle=\color{black},
    commentstyle=\color{gray},
    stringstyle=\color{black},
    showstringspaces=false,
    tabsize=2,
    rulecolor=\color{black},
    framesep=6pt, 
    columns=fullflexible,
}
\title{Language Model as Planner and Formalizer under Constraints}

\author{Cassie Huang\drexel \enspace Stuti Mohan\drexel \enspace Ziyi Yang\brown \enspace Stefanie Tellex\brown \enspace Li Zhang\drexel \\
  \drexel Drexel University \hspace{4pt} \brown Brown University \\
  {\tt \{Cassie.Huang|Harry.Zhang\}@drexel.edu}
}

\begin{document}
\maketitle
\begin{abstract}
LLMs have been widely used in planning, either as planners to generate action sequences end-to-end, or as formalizers to represent the planning domain and problem in a formal language that can derive plans deterministically. However, both lines of work rely on standard benchmarks that include only generic and simplistic environmental specifications, leading to potential overestimation of the planning ability of LLMs and safety concerns in downstream tasks. We bridge this gap by augmenting widely used planning benchmarks with manually annotated, fine-grained, and rich natural language constraints spanning four formally defined categories. Over 4 state-of-the-art reasoning LLMs, 4 formal languages, and 4 datasets, we show that the introduction of one-sentence constraints consistently halves performance, indicating current LLMs' lack of robustness and an avenue for future research.
\footnote{Our code and data can be found at \url{https://github.com/CassieHuang22/LLM-as-Formalizer-constraints}.}

\end{abstract}

\section{Introduction}

Large language models (LLMs) have garnered attention in recent years for their capabilities in planning domains. An intuitive methodology, \textit{LLM-as-Planner}, directly generates a plan given a description of the domain and the problem in an end-to-end manner. While the algorithmic nature of formal planning tasks significantly challenge LLM planners \cite{valmeekam2024planbench,kambhampati2024llms}, the emergence of reasoning LLMs has shown improved performance \cite{valmeekam2024llmscantplanlrms,huang-zhang-2025-limit} but lacks interpretability and control. Another methodology, \textit{LLM-as-Formalizer}, instead translates the input description into some formal languages \cite{xie2023translating,liu2023llm+,zhang-etal-2024-pddlego,zhang-etal-2024-proc2pddl,zhu2024language}. This formal representation can then be passed into a solver which then outputs a plan deterministically, providing more interpretability, control, and promising performance.

\begin{figure*}[t!]
    \includegraphics[width=1\linewidth]{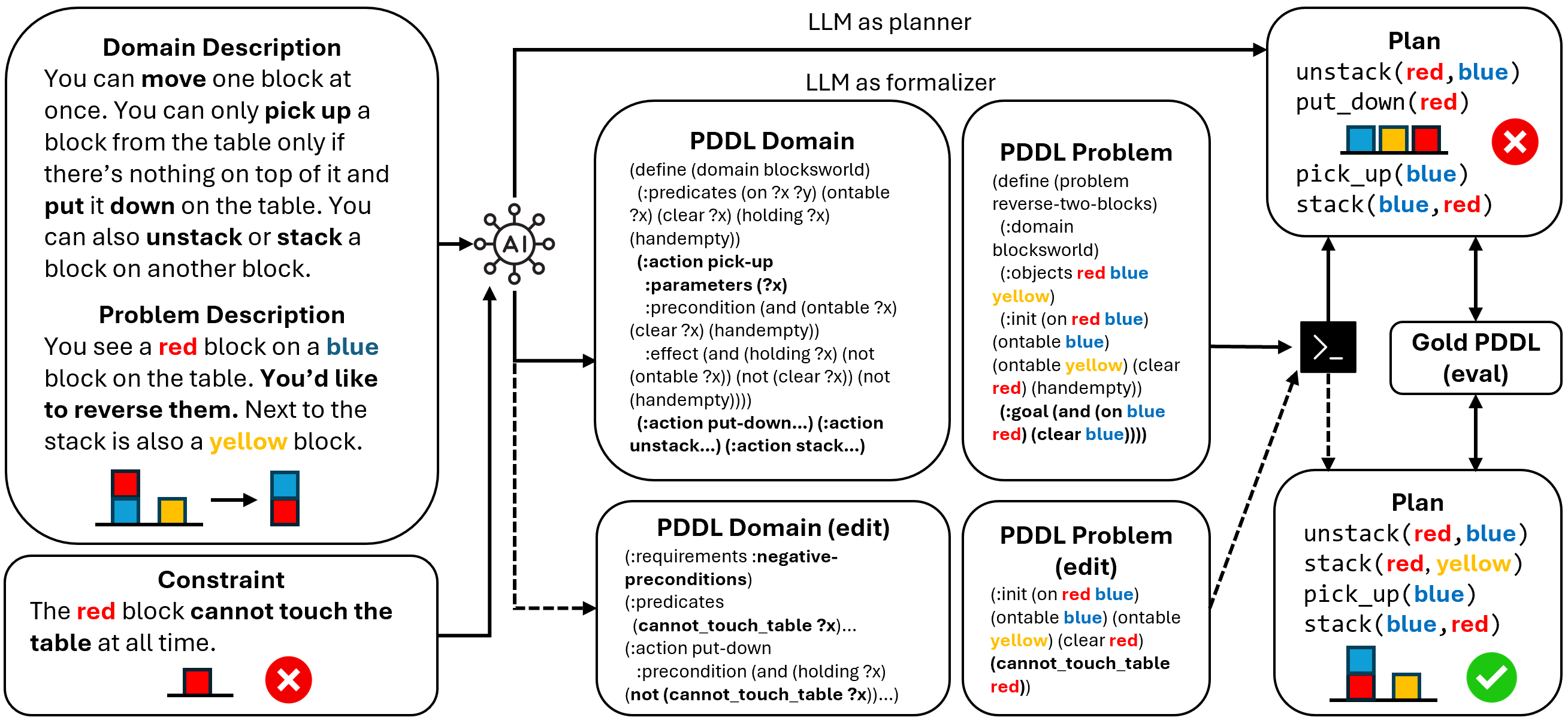}
    \caption{The flow of evaluation on \cope. An LLM is given a description of the planning domain and problem, as well as a natural language constraint. As a planner, the LLM directly outputs a plan consisting of a sequence of actions. As a formalizer, the LLM attempts to construct a formal representation that can invoke an external solver to derive a plan. Alternatively, to address the constraint, the LLM as an editing formalizer (dashed arrows) would first predict PDDL without the constraint before generating edits to it.}
    \label{fig:figure1}
\end{figure*}

As a recent, nascent research direction, most previous work relies on standard planning benchmarks, many of which have existed for decades, where environments are described in a generic, homogeneous, simplistic manner. This is a critical issue because the simplicity of the data may lead to overestimation of the planning or formalizing ability of an LLM. While existing benchmarks enable proof of concept, they are distant from real-life planning instructions that commonly include idiosyncratic requirements inflicted by the user or the resources.

To bridge this gap, we introduce \cope (\textbf{Co}nstrained \textbf{P}lanning \textbf{E}nvironments), a benchmark that augments widely used planning environments such as BlocksWorld \cite{ipc} and CoinCollector \cite{yuan2019countingexploregeneralizetextbased} with \textbf{constraints}. Each example originally includes a domain description, a problem description, and ground-truth PDDL as a simulator to validate a predicted plan. In \cope, we additionally annotate a natural language constraint that is not only linguistically rich but also formally categorized (Figure~\ref{fig:figure1}). 
We systematically evaluate both LLM-as-Planner and LLM-as-Formalizer methodologies, the latter including 4 formal languages including Planning Definition Language (PDDL) 1.2 and 3, Linear Temporal Logic (LTL), and Satisfiability Modulo Theories (SMT). Among 4 state-of-the-art LLMs on 4 datasets, we show that one-sentence constraints consistently degrade performance, indicating lack of robustness of both LLM planning methodologies. To inspire future work, we show that different formalisms respond differently to categories of constraints and also domains. As we corroborate past work by showing the robustness of LLM-as-Formalizer over LLM-as-Planner to problem complexity and lexical shift, we clearly demonstrate that the introduction of constraints substantially negates such robustness. We pose our findings as a timely reflection among a recent slew of methodological work in LLM-based planning, highlighting limitations and opening up future directions. 

\section{Related Work}
\label{sec:related}

As comprehensive literature reviews of the LLM-as-Planner and LLM-as-Formalizer methodology have been accomplished by recent surveys \cite{tantakoun-etal-2025-llms, wei-etal-2025-plangenllms}, we will instead focus on the benchmarks used in existing work. 

\paragraph{Planning Benchmarks} Most work on planning with LLMs experimented with International Planning Competition (IPC) domains including BlocksWorld, Gripper, Logistics, Barman, etc. \cite{kagitha2025addressingchallengesplanninglanguage, hu2025text2worldbenchmarkinglargelanguage, zhu2024language, liu2023llm+, shirai2024visionlanguageinterpreterrobottask, hao2025planning, guo2024castlconstraintsspecificationsllm, shojaee2025illusionthinkingunderstandingstrengths, stein2025automatinggenerationpromptsllmbased, stechly2024chain, valmeekam2024planbench, yao2025spinbenchllmsplanstrategically, guan2023leveraging}. We provide a table of works and a non-exhaustive list of the domains used, as well as examples from common domains in Appendix~\ref{app: benchmark examples}. Beyond those domains and datasets that have existed for decades, few planning benchmarks have been constructed in recent years. Moreover, \citet{zuo2024planetarium,huang-zhang-2025-limit} recently pointed out the templated nature and linguistic homogeneity of these datasets and attempted to describe the domains and problems in a natural fashion. While this modification adds difficulty to the task, the semantics of the domains and problems remain standard, subject to LLMs' memorization, and does not reflect the complexity of real-world planning.

\paragraph{Data Augmentation} Much recent work augmented datasets for different tasks in order to make the problem more realistic, challenging, and less prone to data contamination. This has been done for question answering \cite{Anantheswaran2024CuttingTT, Hong2024TowardsBA}, common sense reasoning \cite{Sun2024EvoGradAD}, instruction following \cite{Hong2024TowardsBA}, and so on. For planning, datasets have been augmented to increase robustness, for example by adding noise \cite{zheng2025diffusionbased} and lexical perturbation \cite{valmeekam2024planbench,zheng2024naturalplanbenchmarkingllms}, with no change to the actual semantics. In contrast, our proposed \cope semantically modifies both the domain and the problems, leading to increased realism and challenge.

\paragraph{Planning With Constraints} In the formal planning community, syntax that expresses some constraints were introduced to the PDDL3 language \cite{gerevini2005plan,10.5555/3037104.3037158}. However, few follow-up work in the NLP and AI community recognized and addressed the real-life challenge of planning with constraints. \citet{yang2023plugsafetychipenforcing} introduced a hybrid system that translates natural language constraints into LTL to assist LLM-as-Planner methods, though their focus was primarily on robotic applications and lacked a thorough discussion of diverse constraint patterns and formal representations. \citet{guo2024castlconstraintsspecificationsllm} encoded constraints with code to interface a SMT solver and combined it with PDDL for task and motion planning though without any public data or code. Moreover, they only considered coarse-grained constraints and oversimplified the task by assuming parts of the PDDL which is often not realistic. We closely consider these formalisms when defining the constraints. 

\section{Task Formulation}
\label{sec:formulation}

\paragraph{Definition of Planning}

We will first formally define the task of text-based formal planning in line with classical STRIPS formalism \citep{10.5555/1622876.1622939} and recent conventions in the natural language processing community \citep{huang-zhang-2025-limit}. 
The input $I$ consists of a triplet $(D_d, D_p, \mathcal{DF'})$:
\begin{enumerate}
    \item $D_d$ is a textual description of the domain, specifying the types (the type system for entities $e_i$), predicates (in relation to the types), and actions $a_i$ including pre-conditions and effects (parameterized by conjunctive logical formulas over the predicates).
    \item $D_p$ is a textual description of the problem, specifying the objects, initial states and goal states in terms of predicates;
    \item $\mathcal{DF_G'}$ is the header of the ground-truth PDDL domain file $\mathcal{DF_G}$ that only provides the names and parameters of the actions. This is required so that the eventual plan is grounded to an existing ontology and can be evaluated fairly.
\end{enumerate}

The final objective is a plan $\pi = [a_0(\bar{e}_0), \ldots, a_m(\bar{e}_m)]$, where each 
$a_i \in A$ is an action provided by $\mathcal{DF_G'}$, and 
each $\bar{e}_i$ is a tuple of grounded entities corresponding to the parameters of $a_i$. 
The plan $\pi$, when executed in the environment expressed by a ground-truth $\mathcal{DF_G}$ and $\mathcal{PF_G}$, must (i) fulfill the pre-conditions of each action, and (ii) constitutes a successful transition from the initial state to the goal state. We will revisit the formalism of plan execution after defining the domain and problem files.

The domain file $\mathcal{DF}$ is defined by three main components:
types ($T$), 
predicates ($R$), and 
action semantics ($a^\textit{param}, \psi_a^\textit{pre},\psi_a^\textit{eff}\,$):
\begin{enumerate}
    \item Types: $T = {t_1, \ldots, t_n}$ is the set of entity types present in the environment.
    \item Predicates: $R = {r_1, \ldots, r_n}$ is the set of relational predicates, each in the form of $r(\bar{t})$, where $\bar{t}$ is a tuple of types that participate in relation $r$.
    \item Action semantics: given an action, the parameters $a^\textit{param}$ are a set of types $\bar{t}$ that participate in this action. The pre-conditions $\psi_a^\textit{pre}: \mathcal{S} \rightarrow \mathbb{B}$ is a boolean formula over the current state which presides over a set of predicates $\bar{r}$, where $S$ is the state space. The action is executable iff $\psi_a^\textit{pre} = \text{True}$. The effects $\psi_a^\textit{eff}: \mathcal{S} \rightarrow \mathcal{S}$ is a formula that transition the current state to the next state after the execution of this action.
\end{enumerate}

The problem file $\mathcal{PF}$ is defined by three main components: 
objects ($E$), 
the initial state ($s_0$), and 
the goal states ($\mathcal{S}_g$):
\begin{enumerate}
    \item Objects: $E = {e_1, \ldots, e_n}$ is the set of named entities present in the environment. Each object is an instance of an entity type $t$ defined in $\mathcal{DF_G'}$.
    \item Initial State: $s_0 \in \mathcal{S}$ is a state defined by set of relational facts of the form $r(\bar{e})$, where $r \in R$ is a relational predicate; $\bar{e}$ is a tuple of entities from $E$ that participate in relation $r$. 
    \item Goal States: $\mathcal{S}_g \subseteq \mathcal{S}$ specifies the set of goal states to be reached. A state $s$ is a goal state if and only if $s \in \mathcal{S}_g$.
\end{enumerate}

Here, the state space $\mathcal{S}$ comprises all possible configurations of relational facts over the object set $E$ with the predicates $R$. 
Each state $s \in \mathcal{S}$ is a set of instantiated facts $\bar{r}(\bar{e})$, describing which relations hold among the objects.
Successfully executing a plan $\pi=[a_0(\bar{e}_0), \ldots, a_m(\bar{e}_m)]$ from the initial state yields a trace of states  $L=[s_0,s_1, \ldots, s_{m+1}]$, where $\psi_{a_i}^\textit{pre} (s_i)=\text{True}$, $s_{i+1}=\psi_{a_i}^\textit{eff} (s_i)$, and $s_{m+1} \in \mathcal{S}_g$.

While LLM-as-Planner maps the input triplet $I$ directly to a plan $L$, LLM-as-Formalizer generates the domain file $\mathcal{DF_P}$ and $\mathcal{PF_P}$ before inputting them into a PDDL solver to search for a plan. 

\begin{table*}
    \centering
    \small
    \begin{tabular}{llll}
    \toprule
         Category & Example constraint  & Formal explanation & Plan \\ \midrule
         None &  &  & \makecell[tl]{unstack(red,blue)\\put\_down(red)\\ pick\_up(blue)\\stack(blue,red)}\\ \midrule
         Initial & \makecell[tl]{The red, blue and yellow blocks are on the table.} & \makecell[tl]{$s'_0=\{\text{on-table(red)}\, \wedge$\\$\text{on-table(blue)}\, \wedge $\\$\text{on-table(yellow)}\}$} & \makecell[tl]{pick\_up(blue)\\stack(blue,red)}\\\midrule
         Goal & \makecell[tl]{You'd like to have all three blocks on the table.} & \makecell[tl]{$\mathcal{S}'_g \models\{\text{on-table(red)}\, \wedge$\\$\text{on-table(blue)}\, \wedge$\\$\text{on-table(yellow)}\}$} & \makecell[tl]{unstack(red,blue)\\put\_down(red)}\\\midrule
         Action & \makecell[tl]{If you move the red block you must move the \\yellow block immediately after.}& \makecell[tl]{$\Pi-\Pi'=\{L \,| \, \exists \, a_i \in 
         L,$\\$ a_i=\text{move(red)}\wedge $\\$a_{i+1}\neq\text{move}(\text{yellow}),$\\$L \in \mathcal{L} \}$}& \makecell[tl]{unstack(red,blue)\\put\_down(red)\\pick\_up(yellow)\\put\_down(yellow)\\pick\_up(blue)\\stack(blue,red)}\\\midrule
         State & \makecell[tl]{Once you start, the number of blocks in each \\stack should not exceed 1 at any time.}& \makecell[tl]{$\mathcal{L}-\mathcal{L}'=\{s \mid \forall \, b\in \overline{b}, $\\$\text{height(b)}>1, s \in \mathcal{S}\}$}& \makecell[tl]{Plan does not exist}\\
        \bottomrule
    \end{tabular}
    \caption{Examples of constraints in each category, explanation of the assignment, and how they would affect the plan of an example problem in BlocksWorld (Figure~\ref{fig:figure1}): ``You see a red block on a blue block on the table. You'd like to reverse them. Next to the stack is also a yellow block.''}
    \label{tab:constraint examples}
\end{table*}

\paragraph{Definition of Constraint}

While the definition of \textit{constraint} varies by community, we take a linguistic and pragmatic approach and define it as supplementary, non-destructive information to the task description that restricts possible behaviors in a plan. We aim to power real-world systems that can adhere to any constraint expressed by human users, encompassing formal definitions from established fields (e.g., classical planning) without being limited by them. Specifically, \cope includes formally categorized constraints as follows. 

We consider a short text $\mathcal{C}$ as a constraint when a plan is valid if and only if it satisfies each of the conditions specified by $\mathcal{C}$.
We introduce two concepts \emph{primitive} and \emph{modified}, where applying constraints modifies the primitive action or state space.
\begin{definition}
    An action is a \textbf{primitive action} $a^*$ of an action $a$, if $\psi_{a}^\textit{pre} \models \psi_{a^*}^\textit{pre}, \psi_{a}^\textit{eff} \models \psi_{a^*}^\textit{eff}$.
\end{definition}
\begin{definition}
    A state is a \textbf{primitive state} $s^*$ of a state $s$, if $s \models s^*$.
\end{definition}
Assuming constraints only introduce new predicates but do not remove any existing ones, every modified action will find exactly one primitive action, and every modified state will find exactly one primitive state from the original action and state space, respectively. 
We then define the constraint types as follows:

\begin{definition}
    Let $L'^*=[s'^*_0, \ldots, s'^*_T]$ denote a trace of primitive states after introducing $\mathcal{C}$, and $s^*_0$ be initial states without $\mathcal{C}$, $\mathcal{C}$ is an \textbf{initial constraint} iff $s'^*_0 = s'_0$ (no new predicate is introduced to the initial state) and $s'^*_0 \neq s^*_0$ (the initial state changes after introducing $\mathcal{C}$).
\end{definition}

\begin{definition}
    Let $\mathcal{S}_g$ and $\mathcal{S}'_g$ be goal states before and after introducing $\mathcal{C}$, $\mathcal{C}$ is a \textbf{goal constraint} iff $\{s \mid s \in \mathcal{S}_g, s \in \mathcal{S}\} \neq \{s'^* \mid s' \in \mathcal{S}'_g, s' \in \mathcal{S}'\}$.
\end{definition}

\begin{definition}
    Let $\pi'^*=[a'^*_0, \ldots, a'^*_T]$ denote a sequence of primitive actions after introducing $\mathcal{C}$, and $\Pi$ and $\Pi'$ be all valid plans before and after introducing $\mathcal{C}$, $\mathcal{C}$ is an \textbf{action constraint} iff $\,\Pi'^* \subsetneq \Pi^*$, where $\Pi'^*$ is a collection of all $\pi'^*$, $\pi' \in \Pi'$.
\end{definition}

\begin{definition}
    Let $L'=[s'^*_0, \ldots, s'^*_T]$ denote a trace of primitive states after introducing $\mathcal{C}$, and $\mathcal{L}$ and $\mathcal{L}'$ be all valid state traces before and after introducing $\mathcal{C}$, $\mathcal{C}$ is a \textbf{state constraint} iff $\mathcal{L}'^* \subsetneq \mathcal{L}$, where $\mathcal{L}'^*$ is a collection of all $L'^*$, $L' \in \mathcal{L}'$, and $\mathcal{C}$ is not any other constraint defined above.
\end{definition}
Our categorization is complete as the definition of state constraints subsumes all possible constraints. The proof of completeness can be found in Appendix \ref{app: completeness proof}. By design, each category of constraints is expressed differently for each formalism (PDDL, PDDL3, LTL, SMT), requiring different level of modification to the original code. 
Notably, the union of the four categories is not trivialized by any formalism. For example, simply using the \textit{constraint} syntax in PDDL3 cannot express action constraints or state constraints requiring invention of new predicates. 
While conjunction of multiple predicates is possible, we do not discuss it in this work. Table~\ref{tab:constraint examples} shows examples of constraints and justification of their categories.

\section{Data}

To evaluate LLM planning in environments with constraints, we propose the \cope benchmark that includes two widely used planning domains:

\noindent\textbf{BlocksWorld} \cite{ipc} is a domain to rearrange stacks of blocks on a table using a robotic arm. It involves four actions: \textit{pick up}, \textit{put down}, \textit{stack}, and \textit{unstack}.
We build on the \bw benchmark from \citet{huang-zhang-2025-limit} with 100 problems, each including a domain and problem description as well as the headers of actions $(D_d, D_p, \mathcal{DF'})$ as input to models. It also comes with ground-truth PDDL $\mathcal{DF}$ and $\mathcal{PF}$, which are used to validate a predicted plan. 


\noindent\textbf{CoinCollector} \cite{yuan2019countingexploregeneralizetextbased} is a domain to navigate a house full of rooms in order to find a coin. It involves two actions: \textit{move} and \textit{pick up}.
Originally designed to evaluate interactive agents, this is a partially observable environment where no complete information is given and so making a complete plan is impossible. To place it under the framework of \cope, we convert it into a fully observable environment. This is done by using depth-first search, and finding all room arrangements to create problem descriptions and ground-truth problem files. We sample 20 problems with 3, 5, 7, 9, and 11 rooms using the benchmark from \citet{jansen2022textworldexpress}. In total, we also have 100 problem instances consisting of input and ground-truth PDDL.

While these two domains are likely included in modern LLMs' training data, they remain standard in literature and meaningfully challenge state-of-the-art LLMs. \cope itself is a step forward to address memorization by changing the semantics of the domains and problems. 
For each domain, we manually annotate 100 constraints spanning the four categories with ground-truth code of each formalism. Data annotation was done with three trained annotators writing non-overlapping constraints before a filtering process. Details can be found in Appendix \ref{app: annotator info}. Our manually annotated constraints are linguistically diverse. Each category of constraints contains different “families” of constraints where the language within each family is similar. For example, “If you move block1, you must move block2 after.” and “If you move block1, you must move block3 after.” are part of the same family for the action category in BlocksWorld. For BlocksWorld, the action constraint category contains 10 different families, state constraints contain 5 families, and initial and goal constraints both contain 4 families of constraints. Across both domains, the average number of words per constraint is 15.885 words. Unlike the original datasets, we allow problems to occasionally have no solutions.
In the full set of \cope, each constraint is paired with each problem, resulting in 10,000 constrained planning problems. To evaluate the most representative cases while keeping analysis tractable, we perform our following evaluation on a subset where we manually pair each constraint and one representative problem, leading to 100 problems per domain.
Examples are shown in Appendix~\ref{sec:data_examples}. 

\begin{figure*}[t!]
    \centering
    \includegraphics[width=\linewidth]{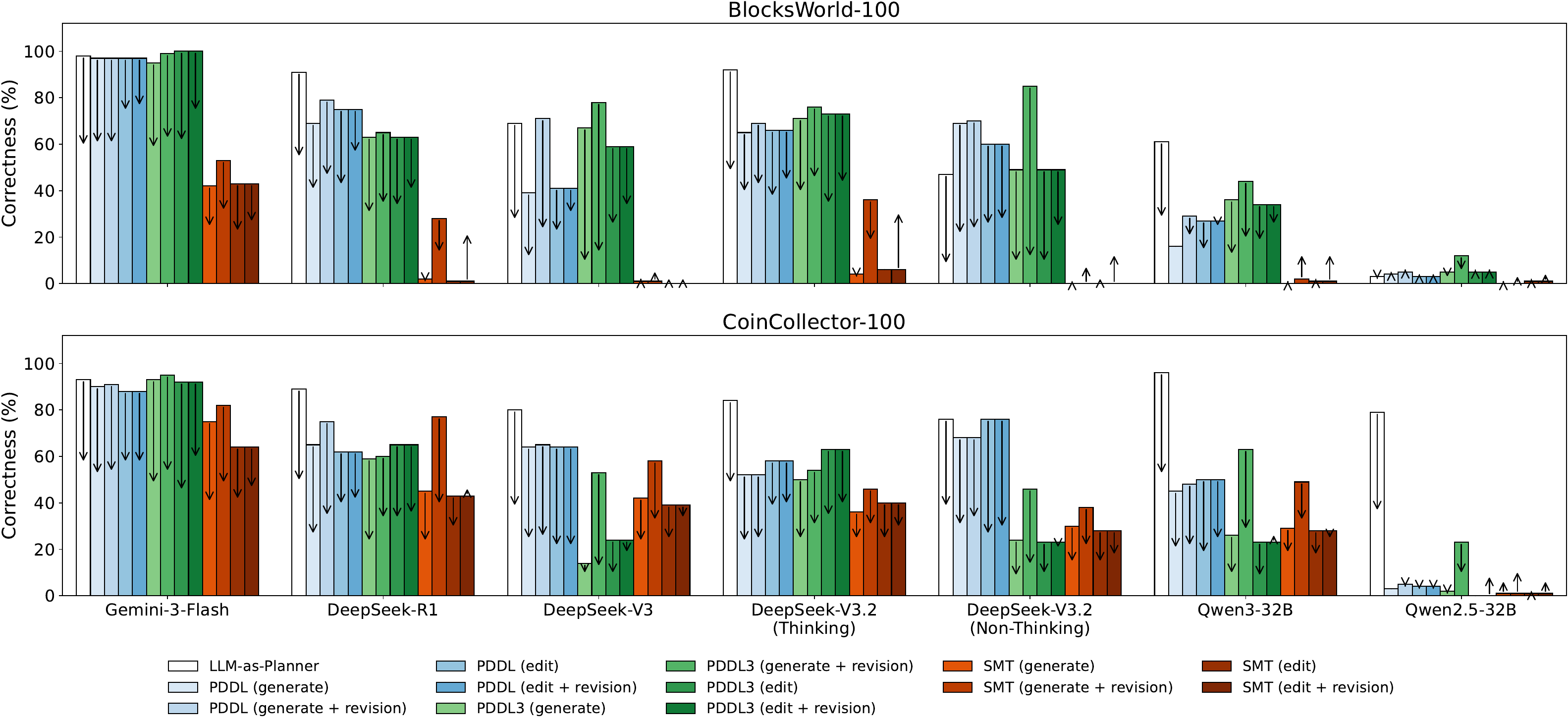}
    \caption{Performance of combinations of LLMs and methods on two domains. The arrows represent the performance difference once constraints are added. Revision is not included in this result to demonstrate that without extra help, this task is difficult. For all results, see Appendix~\ref{app:detailed_results}.}
    \label{fig: constrained vs. non-constrained}
\end{figure*}

\begin{figure*}[!t]
    \centering
    \includegraphics[width=\linewidth]{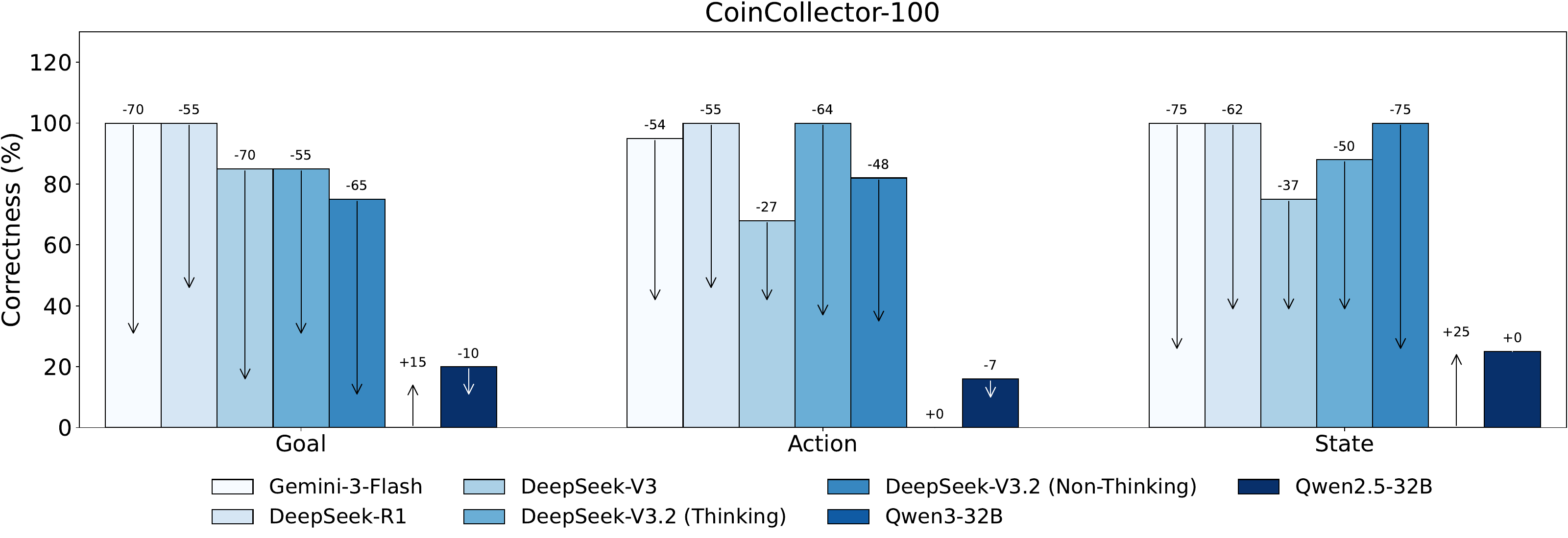}
    \caption{Performance of LTL on a subset of \coin problems. The arrows represent the performance change with constraints.}
    \label{fig: ltl}
\end{figure*}

\section{Experimental Setup}

Given the input with the natural language constraint $(D_d, D_p, \mathcal{DF'}, \mathcal{C})$, we prompt the LLM to generate 4 types of outputs: a plan directly (\textbf{LLM-as-Planner}), or formal programs including \textbf{PDDL} (1.2), \textbf{PDDL3}, \textbf{SMT} (more precisely, code to interact with the Python Z3 library), and \textbf{LTL}. The generation of all output types is done via one-shot prompting following cited work in Section~\ref{sec:related}. When generating formal programs, we consider three techniques:  \textbf{Generation}, where the model directly generates the code in that formal language, \textbf{Editing}, where the model first generates the code without considering the constraint, suggests an edit to that code to address the constraint, and re-generates the code accordingly, and \textbf{Revision}, where the model is given up to 3 tries to revise the code via re-generation based on any error message returned by executing the code, if any. We would like to note that LLM-as-Formalizer was given 3 attempts to write syntactically correct PDDL code, not re-plan. Both methodologies were hence given 1 opportunity to plan. This design choice is fair because a syntax check is almost free of cost, while evaluating a plan is not. 

Following the majority of cited work in Section~\ref{sec:related}, the predicted plan resulting from any method is validated against the ground-truth PDDL provided by \cope, instead of being compared against ``ground-truth'' plans \cite{lyu-etal-2023-faithful,liu2023evaluating,pan-etal-2023-logic} since there could be multiple correct plans. 
We evaluate the predicted plans using the \textit{correctness} metric which indicates the percentage of plans that are successfully validated. For PDDL output, we use the \texttt{dual-bfws-ffparser} planner implemented by \citet{muise-icaps16demo-pd} as the solver. 
For the validator, we use VAL \cite{1374201}. See more details on VAL in Appendix~\ref{sec:solver info}. For PDDL3, the output is first compiled using TCORE \cite{Bonassi_Gerevini_Percassi_Scala_2021} into standard PDDL, then solved and validated the same way as PDDL. For SMT output, the LLM is responsible for generating code that includes a call to invoke the Z3 solver. For LTL, the LLM generates formulas representing environment dynamics, initial and goal states, and constraints. A satisfying plan is then produced by searching for a Lasso-shaped accepting run over the conjunction of all formulas. 
The language model is asked to generate LTL formulas for the initial state, goal state, environment dynamics, and action dynamics. All formulas are concatenated into a conjunctive form as the formal representation for planning. When a constraint is introduced, we make a separate call to the language model for the formula representing the constraint and concatenate the formula to the existing conjunctive formula.
We employ the spot library~\cite{spot-ltl} for encoding LTL formulas into Büchi automatons.
To make the results compatible with our evaluation pipeline, we map the traces back to sequences of PDDL actions by inferring actions from abstract state changes. In this process, the language model is also asked to generate environmental information, which is the room connectivities in CoinCollector domain. We prompt the LLM to generate formulas for goal, action, and state constraints on \coin only, since overriding initial states by editing is practically infeasible when all components of the environment and the problem are represented uniformly in a conjunctive formula. Notably, different from PDDL and SMT, LTL is restricted to domains with simple action dynamics, e.g., navigational domains like CoinCollector, so while it performs well without constraints, it is not as generalizable as other languages.

We use Gemini-3-Flash, DeepSeek-R1, DeepSeek-V3 \cite{guo2025deepseek}, DeepSeek-V3.2 in both Thinking and Non-Thinking mode\footnote{As the endpoints for \texttt{deepseek-reasoner} and \texttt{deepseek-chat} changed in the DeepSeek API on December 1st, 2025, we are confident that experiments for LLM-as-Planner, LLM-as-PDDL-Formalizer, LLM-as-SMT-Formalizer and LLM-as-LTL-Formalizer use the same models. However experiments for LLM-as-PDDL3-Formalizer may have used different models. Therefore we include experiments for DeepSeek-R1 and DeepSeek-V3, as well as the Thinking mode and Non-Thinking mode of DeepSeek-V3.2.} \cite{deepseekai2025deepseekv32pushingfrontieropen}, Qwen3-32B \cite{qwen3technicalreport}, and Qwen2.5-32B \cite{hui2024qwen2} as LLMs that span different sizes of the ability to scale reasoning tokens during inference.
We query Deepseek models with their API and Qwen models using KANI \cite{zhu-etal-2023-kani} with default temperature on 1 H100 GPU. Prompts can be found in Appendix~\ref{sec:prompts}.

\section{Results}
\label{sec:results}

With evaluation on \cope, we answer an array of research questions, draw findings, and provide actionable recommendations for future work.

\begin{figure*}[t!]
    \centering
    \includegraphics[width=\linewidth]{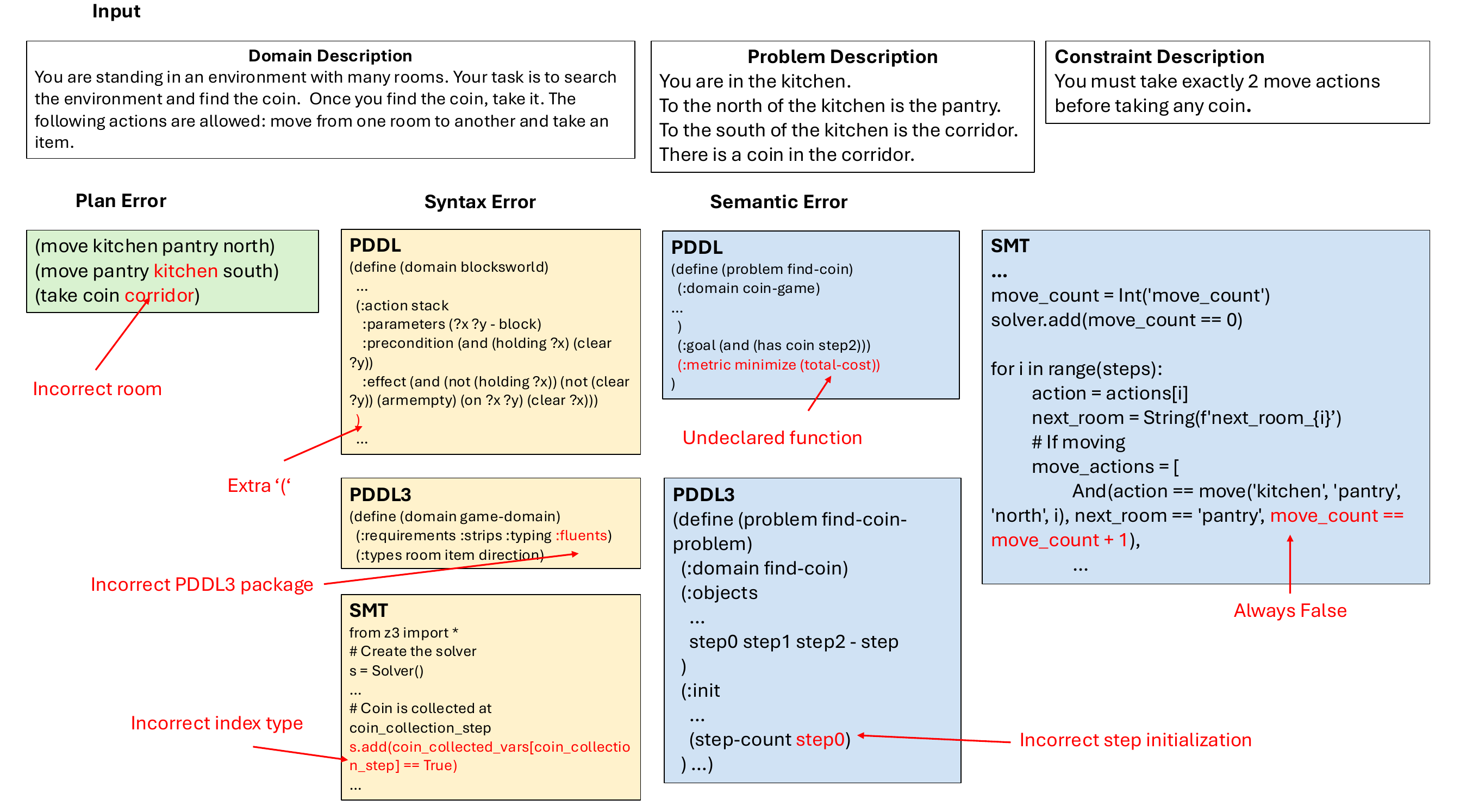}
    \caption{Examples of different errors found in model outputs.}
    \label{fig: error types}
\end{figure*}

\begin{figure*}[t!]
    \centering
    \includegraphics[width=\linewidth]{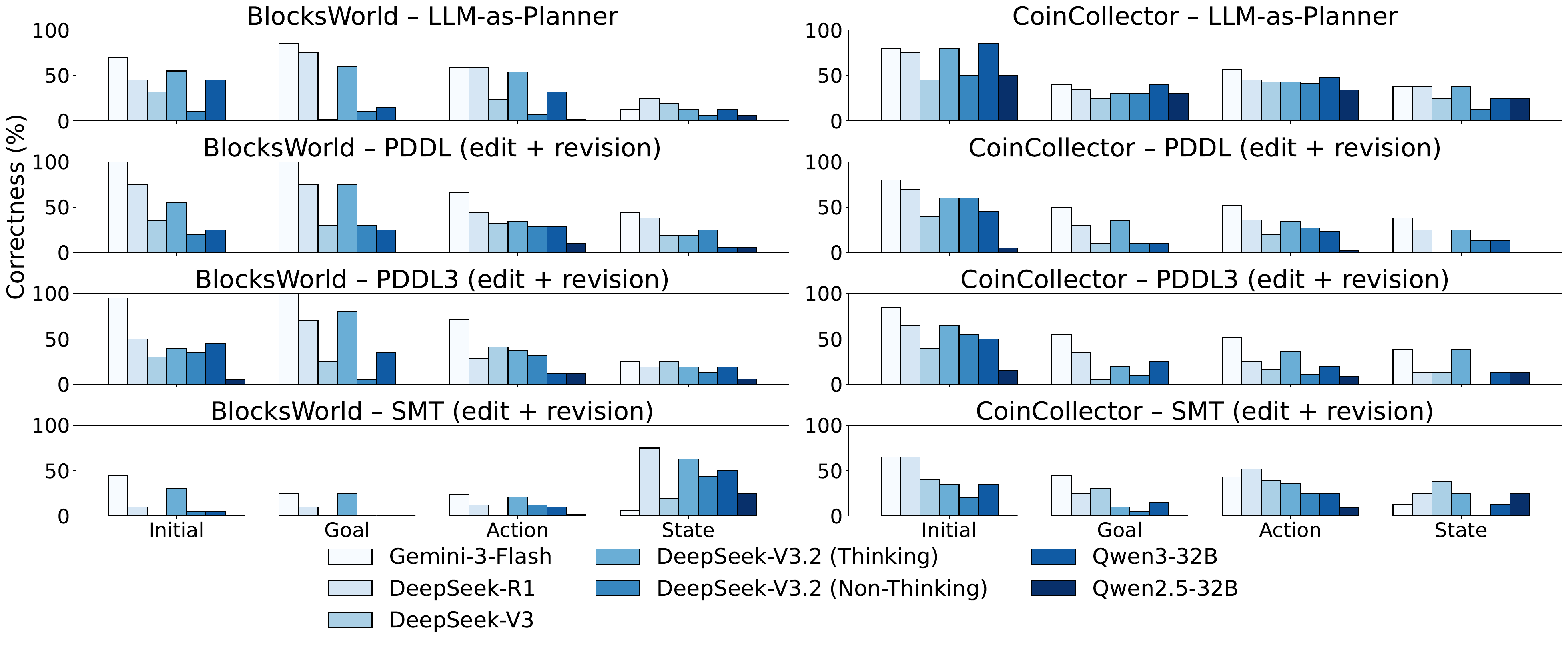}
    \caption{Performance of  LLM-as-Planner and LLM-as-Formalizer (Edit) for PDDL, PDDL3 and SMT by constraint categories on \bw and \coin.}
    \label{fig: constraint category performance}
\end{figure*}

\paragraph{How challenging are the constraints?}
Figure~\ref{fig: constrained vs. non-constrained} and~\ref{fig: ltl} display the performance of all methods, languages, techniques, LLMs, and domains. It is evident that simple one-line \textbf{constraints greatly diminish performance in both planning and formalization}. While all experiments maintain non-trivial correctness greater than 20\%, introducing constraints consistently degrades (often halves) the planning performance over all cases. Even the most powerful model, Gemini-3-Flash sees performance decrease by around 30\%. Overall, LLM-as-Planner outperforms its LLM-as-Formalizer counterparts before and after the introduction of constraints. However, LLM-as-Formalizer, with interpretability and formal guarantee, shows competitive performance with Gemini and larger DeepSeek models as well as using revision by solver errors. With constraints, it benefits from the two-step generate-then-edit technique over simply generating PDDL in some but not all cases. 
Appendix Figure~\ref{fig: syntax vs semantic} shows that once constraints are introduced, the number of syntax errors increases for most methods, especially with PDDL3.
Comparing the two domains, CoinCollector (2 actions) is much simpler than BlocksWorld (4 actions) with regard to LLM-as-Planner and LLM-as-SMT-Formalizer but comparable with regard to other methods. Intuitively, this evidences LLM-as-PDDL-Formalizer's robustness against the complexity by the number of actions over other methods.
Regarding the choice of formal language, PDDL greatly outperforms SMT in most cases, likely because PDDL more closely expresses the planning problem. However, while seemingly a more appropriate language which models constraints, PDDL3 generation is consistently worse than PDDL due to an elevated amount of compilation and syntax errors, likely because PDDL3 is even lower-resource for LLM pre-training. 


\begin{figure*}[t!]
    \centering
    \small
    \includegraphics[width=.9\linewidth]{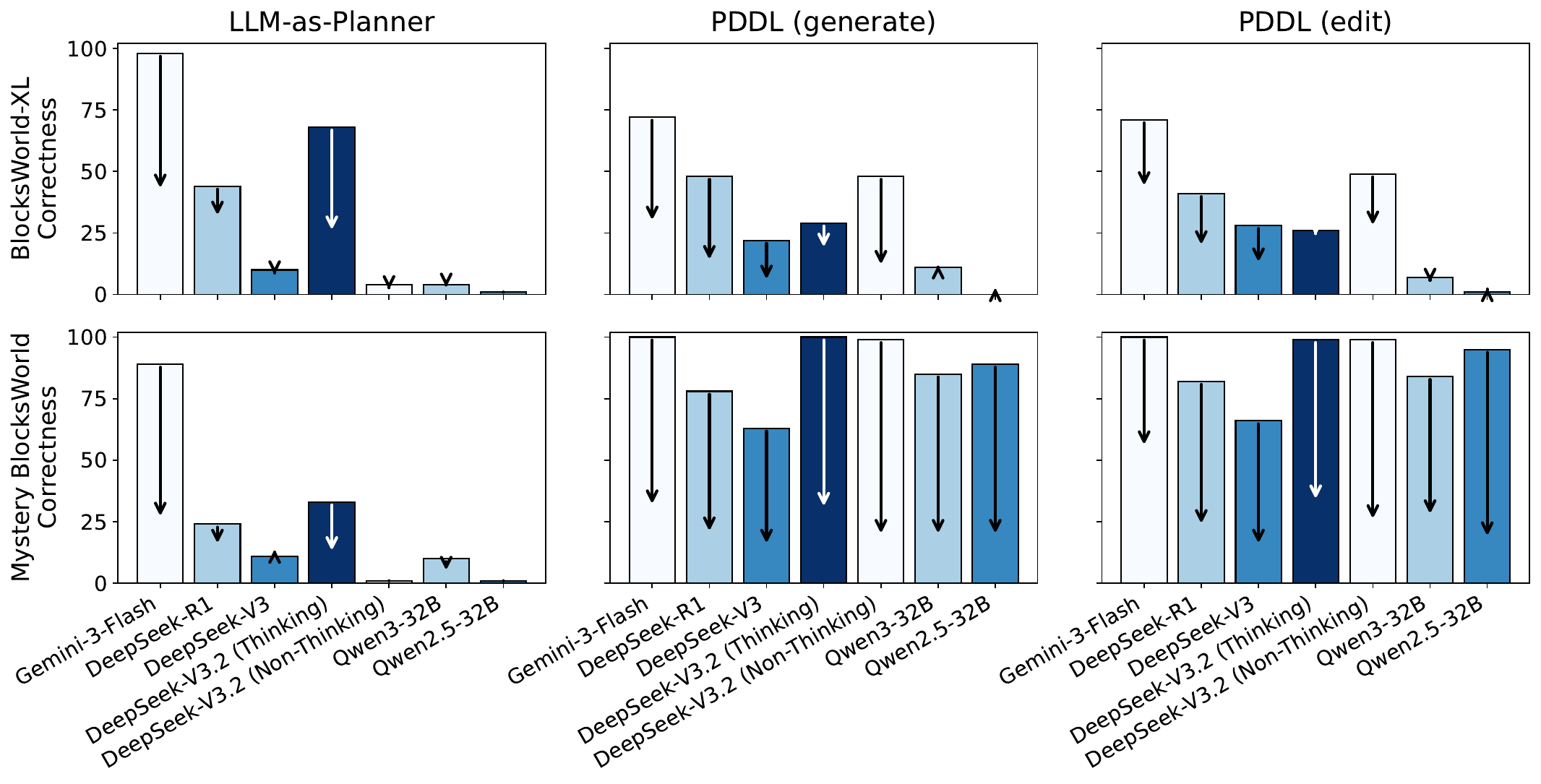}
    \caption{Performance of various methods on \bwxl and \mbw data to study models' robustness to complexity and lexical perturbation respectively, under the influence of constraints. As before, the arrows represent the performance difference once constraints are added. Results for SMT are omitted due to near-zero performance on both datasets.}
    \label{fig: bw-xl and mbw}
\end{figure*}

\paragraph{When do models fail?}
Figure \ref{fig: error types} qualitatively demonstrates an example set of errors for a particular constraint. A single constraint may cause various types of errors. For LLM-as-Planner this set of errors includes hallucinating plans that do not exist, not satisfying newly added constraints and not following preconditions or effects set by the actions in the domain. The set of errors for LLM-as-Formalizer includes syntax errors of PDDL, PDDL3 or Python, declaration of logically-incorrect functions, predicates, action parameters, preconditions in PDDL and PDDL3, or false encoding of constraints in SMT. 
Once revision is introduced, the issues are mainly semantic. Within semantic errors,  the most common errors include incorrect use/missing of the constraints section in PDDL3 and missing/incorrect predicates needed to correctly encode the constraints for PDDL.

\paragraph{What category of constraints is hard?}
One may hypothesize that some categories of constraints may pose more challenge than others for a particular method. Recalling the definitions in Section~\ref{sec:formulation}, local edits may allow LLMs as PDDL formalizers to satisfy initial and goal constraints more easily compared to state constraints, since they do not require introducing and applying new predicates. In contrast, state constraints may be much easier to model using SMT which provides analogous functions. Moreover, PDDL3 may do well on both initial and goal constraints (due to local edits), as well as state constraints since the syntax can now encode constraints. Figure \ref{fig: constraint category performance} shows some yet non-decisive evidence towards those hypotheses. Figure \ref{fig: ltl} demonstrates that state constraints are more difficult than action constraints for LTL using Gemini-3-Flash and DeepSeek models. 
Furthermore, PDDL3 demonstrated similar results to standard PDDL and in contrast to initial assumptions, performed the worst on state constraints.

\paragraph{Can LLMs scale with complexity?}
Recent work has cast doubt on the ability of LLMs, as planner and formalizers, to scale with the problem complexity \cite{shojaee2025illusionthinkingunderstandingstrengths,lin2025zebralogic,amonkar2025llmsbetterformalizerssolvers}. In domains considered in this work, problems from \bw range from 2 to 15 blocks, while those from \coin range from 3 to 11 rooms. Although our results represented in Figure~\ref{fig: constrained vs. non-constrained} pose a significant room for improvement in model performance, especially after the introduction of constraints, it would only be more meaningful to conduct studies on problems with higher complexity emulate more to real life applications. 
Therefore, we present \bwxl, a dataset included in \cope where all problems contain 50 blocks, in order to see the impact of problems with a larger entity space on planning, especially with constraints. 

Comparing performance of the first row of Figure \ref{fig: bw-xl and mbw} to that on \bw (Figure~\ref{fig: constrained vs. non-constrained}), performance for all methods, with the exception of Gemini-3-Flash, sharply decreases as the problems become more complex even without the introduction of constraints, in line with cited literature. 
Even so, LLM-as-Formalizer displays more robustness to complexity than LLM-as-Planner. However, \textbf{such scaling behavior disappears with the constraints} as performance degrades by more than 50\% for the well-performing DeepSeek models. This finding clearly indicates the challenge caused by the constraints aggravates the existing concern of scaling inability in LLM-assisted planning methodologies.

\paragraph{Do LLMs memorize the domains?}
Among the domains included in \cope, BlocksWorld has existed for decades; and the simple semantics of navigation in CoinCollector have existed in many longstanding planning domains. As state-of-the-art LLMs are trained with most publicly available data, data contamination has been a stubborn challenge that hinders research \cite{sainz-etal-2023-nlp}. As we have discussed in Section~\ref{sec:related}, we pose \cope and the introduction of constraints as a remedy to shift the semantics in the data and bring models out of their comfort zone. We take one step further and apply the lexical obfuscation method of Mystery BlocksWorld \cite{valmeekam2024planbench} to \bw to construct \mbw, by replacing all the names of the types, predicates, actions, and objects with nonsensical placeholders. We add our constraints by similarly swapping out the concepts in the constraints and the ground-truth PDDL. 

From the second row of Figure \ref{fig: bw-xl and mbw}, the performance without constraints is consistent with that of \citet{huang-zhang-2025-limit}, where LLM-as-Planner is devastated compared to the original \bw, suggesting some extent of memorization, while LLM-as-Formalizer remains robust to lexical perturbation. However, \textbf{such robustness disappears with the constraints}, as the performance for all 4 LLMs and 2 formalizing methods decreases by two thirds or more. 

Our results jointly show that LLM-based planning and formalization are far from achieving robust performance on complex problems in different domains. They also emphasize the importance of future benchmarks like \cope that transcend simple problems in typical domains.

\section{Conclusion}
We introduce \cope, a benchmark that augments existing planning domains with linguistically rich and formally categorized constraints to increase complexity and real-world applicability. Results show that \cope presents significant challenges and consistently degrades performance. We provide a detailed analysis of behavioral differences of methods and planning languages, suggesting future research on tooling for LLM-based planning. Importantly, our constraints amplify models' lack of robustness with increasing problem complexity and lexical perturbation, calling for attention to the gap between benchmarks and real-world planning. 

\section{Limitation}

Despite our best effort in defining, formalizing, and categorizing the constraints, we inevitably fall short of the goal of addressing all possible constraints (or more generally, requests) expressed in natural language. Even within our formalism, interesting phenomena such as the conjunction, negation, or ambiguity of the constraints are left for future work. 

Our evaluation metric, plan correctness, may induce false positives where the plan may happen to be correct, but the generated code does not actually correctly describe the environment that satisfies the constraint, leading to concerns in interpretability and faithfulness. However, to the best of our knowledge there is no feasible alternative. Nevertheless we perform an analysis on 20 samples from all datasets and methods and there were no false positives. Therefore the false positive rate is negligible. While there exists methods to evaluate generated PDDL problem files \cite{zuo2024planetarium} or domain files \cite{coulter2022theory} against a ground-truth, fundamentally there exists non-unique ways to formulate the same constraint. 

While \cope augments 4 datasets (all under the MIT License) in the BlocksWorld and CoinCollector domains, they are still oversimplified proof of concept and not representative to problems in the real world. This may pose a risk when applied to real-world problems. Still, the challenge to state-of-the-method as we have demonstrated argues for the necessity of incremental evaluation.

The deployment of autonomous agents in downstream tasks introduces significant safety risks that vary across domains. In embodied systems, such as household robots, poorly specified constraints can lead to catastrophic physical harm \cite{yang2023plugsafetychipenforcing}. Similarly, AI agents in virtual domains can be exploited to execute complex harmful sequences, such as facilitating self-harm \cite{zhang2025agentalignnavigatingsafetyalignment}. These failure modes exemplify the risks of dual-use and underspecification. Our method mitigates these risks by converting ambiguous natural language into a formal, interpretable representation. This transition provides the transparency necessary for human-in-the-loop auditing and formal verification, ensuring that agent behavior aligns with intended safety constraints rather than just operational commands.

\section*{Acknowledgement}

This research is funded through the National Science Foundation’s Civil Infrastructure Systems (CIS) program, award number 2409847. Any errors or omissions remain the sole responsibility of the authors.

\bibliography{anthology, custom}

\appendix
\section{Benchmark Examples} \label{app: benchmark examples}
Table \ref{tab:related_works} provides related works and a non-exhaustive list of domains used. Listings \ref{lst:moderately_templated_blocksworld_dd} and \ref{lst:moderately_templated_blocksworld_pd} display a benchmark example of BlocksWorld from \citet{valmeekam2024planbench} and \cite{huang-zhang-2025-limit}. Listings \ref{lst:moderately_templated_logistics_dd} and \ref{lst:moderately_templated_logistics_pd} display an example of the Logistics domain from \citet{huang-zhang-2025-limit}. Language seen in this benchmark is simplistic and not realistic to the real world.
\begin{table*}[!t]
    \small
    \centering
    \begin{tabularx}{\textwidth}{l|lX}
    \toprule
         \multicolumn{1}{c}{} & & Domains Used \\ \midrule
         & \citet{huang-zhang-2025-limit} & BlocksWorld, Logistics, Barman\\
         & \citet{kagitha2025addressingchallengesplanninglanguage} & BlocksWorld, Logistics, Barman \\
         & \citet{gong2025zeroshotiterativeformalizationplanning} & CoinCollector, ALFWorld \\
         & \citet{hu2025text2worldbenchmarkinglargelanguage} & Over 100 domains including Grid and BlocksWorld \\
         & \citet{li2024embodied} & BEHAVIOR, VirtualHome \\
         & \citet{zuo2024planetarium} & BlocksWorld, Gripper \\
         & \citet{zhu2024language} & Barman, BlocksWorld, Floortile, Grippers, Storage, Termes, TyreWorld \\
         & \citet{zhang-etal-2024-pddlego} & CoinCollector, CookingWorld \\
         & \citet{zhang-etal-2024-proc2pddl} & Proc2PDDL \\
         & \citet{liu2023llm+} & Barman, BlocksWorld, Floortile, Grippers, Storage, Termes, TyreWorld \\
         LLM-as-Formalizer & \citet{xie2023translating} & BlocksWorld, ALFRED-L \\
         & \citet{wong2023learning} & Mini Minecraft, ALFRED \\
         & \citet{guan2023leveraging} & Household, Logistics, TyreWorld \\
         & \citet{shirai2024visionlanguageinterpreterrobottask} & Cooking, BlocksWorld, Tower of Hanoi \\
         & \citet{ishay2025llmalbridginglargelanguage} & MCP \\
         & \citet{hao2025planning} & Coffee, Workforce, Facility, Task Allocation, Warehouse, BlocksWorld, Mystery BlocksWorld, Movie, Gripper \\
         & \citet{guo2024castlconstraintsspecificationsllm} & HouseChip, Kitchen, BlocksWorld \\
         & \citet{tang2024worldcodermodelbasedllmagent} & Sokoban, Minigrid, ALFWorld \\
         & \citet{wang2023bytesized32corpuschallengetask} & ByteSized32 \\
         & \citet{amonkar2025llmsbetterformalizerssolvers} & NaturalPlan, ZebraLogic \\
         & \citet{lyu-etal-2023-faithful} & SayCan \\
         & \citet{gao2023pal} & GSM8K \\
         \midrule
         & \citet{shojaee2025illusionthinkingunderstandingstrengths} & Tower of Hanoi, Checker Jumping, BlocksWorld, River Crossing \\
         & \citet{parmar2025plangenmultiagentframeworkgenerating} & NaturalPlan, GPQA, OlympiadBench \\
         & \citet{majumder2023clincontinuallylearninglanguage} & ScienceWorld \\
         & \citet{lin2023swiftsage} & ScienceWorld \\
         LLM-as-Planner & \citet{stein2025automatinggenerationpromptsllmbased} & BlocksWorld, Depot, Logistics, Gripper, Ferry, Floortile, Goldminer, Grid, Gripper, Movie, Rovers, Satellite, Visitall \\
         & \citet{zheng2024naturalplanbenchmarkingllms} & NaturalPlan \\
         & \citet{stechly2024chain} & BlocksWorld \\
         & \citet{valmeekam2024planbench} & BlocksWorld, Logistics, Mystery BlocksWorld \\
         & \citet{yao2025spinbenchllmsplanstrategically} & BlocksWorld, Gripper, Barman, Logistics and more \\
         
\bottomrule
    \end{tabularx}
    \caption{Comparison with related work.}
    \label{tab:related_works}
\end{table*}
\section{Data Examples}
\label{sec:data_examples}

We augment constraint descriptions to a ground-truth domain and problem file and create new ground-truth PDDL files that now encode the ground-truth. Here we include an example constraint description and the ground-truth PDDL files for the constraint + problem pair. Listings \ref{lst:constraint_description}, \ref{lst:blocksworld_df} and \ref{lst:blocksworld-pf} display an example constraint description and the annotated ground-truth \df and \pf for a paired BlocksWorld problem. Listings \ref{lst:coin-collector_description}, \ref{lst:coin-collector_df} and \ref{lst: coin-collector_pf} display an example Domain, Problem and Constraint Description for \coin as well as their groundtruth \df and \pf in PDDL. Listing \ref{lst:blocksworld-pf-xl} shows an example ground-truth \pf for the XL dataset, the \df remains the same. Listings \ref{lst:constraint_description-mbw}, \ref{lst:mystery_blocksworld-df} and \ref{lst:mystery_blocksworld-pf} display the corresponding constraint description, \df and \pf that have now been obfuscated.

\section{Annotation Information} \label{app: annotator info}
All annotators were authors of this paper and were asked to annotate the PDDL given the problem description and constraint description. Since no crowdworkers or external annotators were involved, considerations about fair monetary compensation relative to geographic location do not apply. Since no data was collected from human subjects and no personal information is present, consent procedures are not applicable. This study does not constitute human subjects research and did not require ethics review board approval.

\section{Prompts}
\label{sec:prompts}

Listing \ref{lst:prompt_LLM-as-Planner} is the prompt for LLM-as-Planner. Listings \ref{lst:prompt_LLM-as-Formalizer-generate}, \ref{lst:prompt_LLM-as-Formalizer-edit1}, \ref{lst:prompt_LLM-as-Formalizer-edit2} and \ref{lst:prompt_LLM-as-Formalizer-revision} display the prompts for all experiment settings given to all models for LLM-as-PDDL-Formalizer. Listings \ref{lst:prompt_pddl3-generate}, \ref{lst:prompt_pddl3-edit1}, \ref{lst:prompt_pddl3-edit2} and \ref{lst:prompt_pddl3-revision} display the prompts for all experiment settings given to all models for LLM-as-PDDL3-Formalizer. Listings \ref{lst:prompt_llm-as-smt-formalizer}, \ref{lst:prompt_llm-as-smt-formalizer_edit} and \ref{lst:prompt_llm-as-smt-formalizer_revision} display prompts for experiments using SMT. Listing \ref{lst:prompt_llm-as-ltl-formalizer} displays all prompts for LTL. For PDDL and SMT, we asked the model to return the output in a JSON object whenever possible for easier parsing.

\section{Experimental Setup Details} \label{sec:solver info}

\subsection{VAL}
The VAL library takes in a ground-truth PDDL \df, \pf and a plan and tries to execute the plan in the environment by checking whether each action in the found plan can be executed based on the preconditions written in the ground-truth files. If the plan is executable, VAL then checks whether the final state after executing the plan matches the goal state found in the ground-truth \pf. If either the plan is not executable or the final state does not match the goal state, VAL will return an error, therefore the plan found by the planner is not correct.

\section{Completeness of Constraint Type Definitions} \label{app: completeness proof}

\begin{definition}
\label{def:primitive_state}
    A state $s^*$ is a \textbf{primitive state} of a state $s$ if and only if $s \models s^*$. 
\end{definition}

\begin{definition}
\label{def:state_constraint}
    Let $\mathcal{L}$ and $\mathcal{L}'$ be the sets of all valid state traces before and after introducing $\mathcal{C}$, and $\mathcal{L}^*$, $\mathcal{L}'^*$ be their respective primitive state traces. $\mathcal{C}$ is a \textbf{state constraint} iff $\mathcal{L}'^* \subsetneq \mathcal{L}$ and $\mathcal{C}$ is not an initial or goal constraint.
\end{definition}

\begin{theorem}
The set of constraint definitions $\mathcal{D} = \{\text{State, Initial, Goal}\}$ is complete. Provided no existing predicates are removed, any valid constraint $\mathcal{C}$ must fall into at least one of these categories.
\end{theorem}

\begin{proof}
Let $\mathcal{L}$ be the set of all valid state traces $L = [s_0, s_1, \dots, s_{m+1}]$ in the original planning problem. Let $\mathcal{L}'$ be the set of traces after introducing a new constraint $\mathcal{C}$, and let $\mathcal{L}^*$ and $\mathcal{L}'^*$ be their respective sets of primitive state traces.

By definition of the original planning space, we have $\mathcal{L} = \mathcal{L}^*$.
Assume for contradiction that there exists a constraint $\mathcal{C}$ that is not a state constraint. By Definition~\ref{def:state_constraint}, this implies the condition $\mathcal{L}'^* \subsetneq \mathcal{L}$ is false\footnote{Note that if $\mathcal{L}'^* = \emptyset$, the condition $\emptyset \subsetneq \mathcal{L}$ holds (assuming $\mathcal{L} \neq \emptyset$). In this instance, the constraint would be categorized as a state constraint that prunes all paths, while it is considered invalid because its destructive nature violates our definition of a constraint.}. We partition the remaining possibilities for $\mathcal{L}'^*$ into three non-overlapping cases:

\paragraph{Case 1: $\mathcal{L} \subseteq \mathcal{L}'^*$} 
In this case, all original plans producing $\mathcal{L}$ remain valid after the introduction of $\mathcal{C}$. This implies that the primitive representations of the initial states and goal conditions remain unchanged:
\[ s_0^* = s_0'^* \quad \text{and} \quad \{s \mid s \in S_g\} = \{s'^* \mid s' \in S'_g\} \]

However, if $\mathcal{L} \subseteq \mathcal{L}'^*$ and $\mathcal{L} = \mathcal{L}^*$, the introduction of $\mathcal{C}$ either leaves the primitive trace set unchanged or expands it. If $\mathcal{L} = \mathcal{L}'^*$, then $\mathcal{C}$ does not restrict the planning space, contradicting the assumption that $\mathcal{C}$ is a \textit{constraint}. If $\mathcal{L} \subsetneq \mathcal{L}'^*$, there must exist a state $s'$ in the new traces that does not belong to any primitive state $s^*$ in the original traces. Formally:

\[ 
\begin{aligned}\exists \, s' \in L', L' \in \mathcal{L}' \text{ such that } s' \not\models s^*, \, \\
\forall s \in L, \forall L \in \mathcal{L}, 
\end{aligned}
\]

\noindent which contradicts Definition~\ref{def:primitive_state} regarding primitive states and the underlying predicate set. 

\paragraph{Case 2: $\mathcal{L}'^* \not\subseteq \mathcal{L}$ and $\mathcal{L}'^* \, \cap \, \mathcal{L} \neq \emptyset$}
In this case, the set $\mathcal{L}'^*$ contains at least one trace $L'^* \in (\mathcal{L}'^* \setminus \mathcal{L})$ that was not valid in the original problem. This implies either the primitive representations of the initial states or of the goal conditions have changed, i.e., $s_0^*\neq s_0'^* \text{ or } \{s \mid s \in S_g\} \neq \{s'^* \mid s' \in S'_g\}$.

\begin{enumerate}
    \item $s_0^* \neq s_0'^*$: The primitive initial state has been altered, classifying $\mathcal{C}$ as an \textbf{initial constraint}.
    \item $\{s \mid s \in S_g\} \neq \{s'^* \mid s' \in S'_g\}$: The primitive goal states have been altered, classifying $\mathcal{C}$ as a \textbf{goal constraint}.
\end{enumerate}

Consequently, $\mathcal{C}$ is subsumed by the existing definitions of initial or goal constraints.

\paragraph{Case 3: $\mathcal{L}'^* \cap \mathcal{L} = \emptyset$ and $\mathcal{L}'^* \neq \emptyset$}
In this case, no original plans remain valid, and the new trace set is non-empty. Following the logic in Case 2, this necessitates a shift in the initial state $s_0^*$ or the goal states $S_g$, again classifying $\mathcal{C}$ as either an \textbf{initial constraint} or a \textbf{goal constraint}.

\paragraph{Conclusion}
Since all logical configurations of $\mathcal{L}'^*$ either satisfy the definition of a state constraint or are classified as initial/goal constraints, the set $\mathcal{D}$ is complete regardless of the complexity of $\mathcal{C}$.

\end{proof}

\section{Full Results} \label{app:detailed_results}
Tables \ref{tab: bw_LLM-as-Planner}, \ref{tab: bw_pddl}, \ref{tab: bw_pddl3} and \ref{tab: bw_smt} display results for all methodologies for \bw. Tables \ref{tab: cc_planner}, \ref{tab: cc_pddl}, \ref{tab: cc_pddl3}, \ref{tab: cc_smt} and \ref{tab: cc_ltl} display results for all methodologies for \coin. Tables \ref{tab: all blocksworld-xl} and \ref{tab: all mbw} display results for \bwxl and \mbw on PDDL.

\begin{figure*}[t!]
    \centering
    \includegraphics[width=\linewidth]{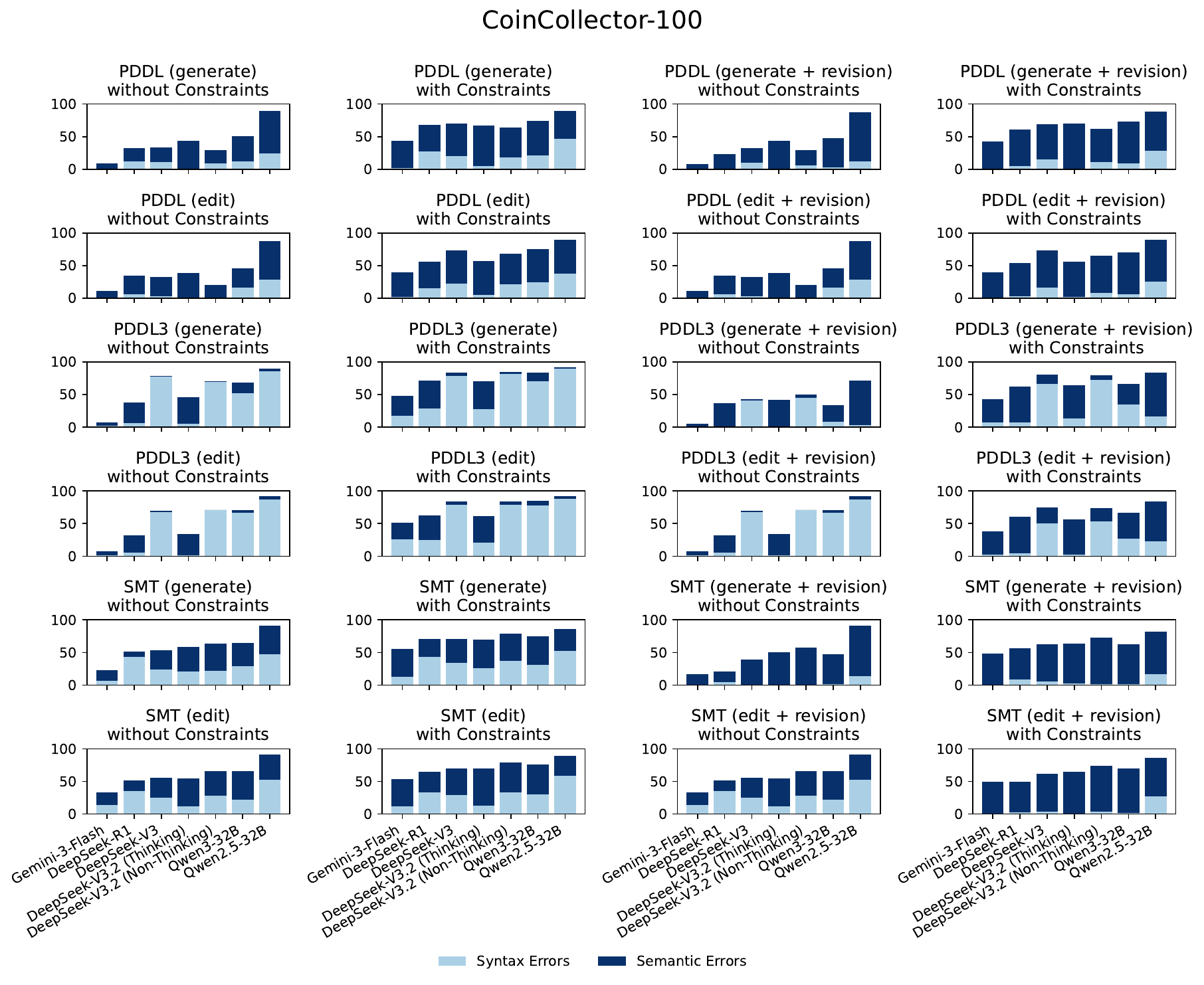}
    \caption{Counts of Syntax vs. Semantic Errors with and without Constraints on \coin using both PDDL and SMT.}
    \label{fig: syntax vs semantic}
\end{figure*}


\clearpage

\onecolumn

\begin{lstlisting}[language=lisp, caption={Domain Description for Moderately Templated BlocksWorld-100}, label={lst:moderately_templated_blocksworld_dd}]
I am playing with a set of blocks where I need to arrange the blocks into stacks. Here are the actions I can do

   Pick up a block
   Unstack a block from on top of another block
   Put down a block
   Stack a block on top of another block
   
   I have the following restrictions on my actions:
   I can only pick up or unstack one block at a time.
   I can only pick up or unstack a block if my hand is empty.
   I can only pick up a block if the block is on the table and the block is clear. A block is clear if the block has no other blocks on top of it and if the block is not picked up.
   I can only unstack a block from on top of another block if the block I am unstacking was really on top of the other block.
   I can only unstack a block from on top of another block if the block I am unstacking is clear.
   Once I pick up or unstack a block, I am holding the block.
   I can only put down a block that I am holding.
   I can only stack a block on top of another block if I am holding the block being stacked.
   I can only stack a block on top of another block if the block onto which I am stacking the block is clear.
   Once I put down or stack a block, my hand becomes empty.
   Once you stack a block on top of a second block, the second block is no longer clear.
    \end{lstlisting}

    \begin{lstlisting}[language=lisp, caption={Problem Description for Moderately Templated BlocksWorld-100}, label={lst:moderately_templated_blocksworld_pd}]
As initial conditions I have that, the blue block is clear, the yellow block is clear, the hand is empty, the blue block is on top of the red block, the yellow block is on top of the green block, the red block is on the table, and the green block is on the table.
My goal is to have that the blue block is on top of the yellow block, the green block is on top of the red block, the yellow block is on top of the green block, and the red block is on the table.
    \end{lstlisting}

\begin{lstlisting}[language=lisp, caption={Domain Description for Moderately Templated Logistics-100}, label={lst:moderately_templated_logistics_dd}]
I need to move packages between locations. Here are the actions I can do

    Load an package onto a truck at a location (load-truck package truck location)
    Load an package onto an airplane at a location (load-airplane package airplane location)
    Unload an package from a truck at a location (unload-truck package truck location)
    Unload an package from an airplane at a location (unload-airplane package airplane location)
    Drive a truck from location1 to location2 in a city (drive-truck truck location1 location2 city)
    Fly an airplane from airport1 to airport2 (fly-airplane airplane airport1 airport2)
    
I have the following restrictions on my actions:
    I can only load a package onto a truck or airplane if both the package and airplane are at the location.
    Once I load the package in the truck or airplane, it is no longer at the location.
    I can only unload a package from a truck or airplane if the truck or airplane is at the location and the package is in the truck or airplane.
    Once I unload the truck or airplane, the object is at the location and no longer in the truck or airplane.
    I can only drive a truck between locations if the truck is at the first location and both the first and second locations are in the same city. Once I drive a truck, the truck is in the second city and no longer in the first city.
    I can only fly an airplane between two airports and the airplane is at the first airport.
    Once I fly an airplane, the airplane is at the second airport and no longer at the first airport.
    \end{lstlisting}

\begin{lstlisting}[language=lisp, caption={Problem Description for Moderately Templated Logistics-100}, label={lst:moderately_templated_logistics_pd}]
As initial conditions, I have that, obj11 is a package, obj12 is a package, obj13 is a package, obj21 is a package, obj22 is a package, obj23 is a package, tru1 is a truck, tru2 is a truck, cit1 is a city, cit2 is a city, pos1 is a location, apt1 is a location, pos2 is a location, apt2 is a location, apt1 is an airport, apt2 is an airport, apn1 is an airplane, apn1 is at apt2, tru1 is at pos1, obj11 is at pos1, obj12 is at pos1, obj13 is at pos1, tru2 is at pos2, obj21 is at pos2, obj22 is at pos2, obj23 is at pos2, pos1 is in cit1, apt1 is in cit1, pos2 is in cit2, and apt2 is in cit2. 
My goal is to have that obj11 is at apt1, obj23 is at pos1, obj13 is at apt1, and obj21 is at pos1. 
    \end{lstlisting}
\begin{lstlisting}[language=lisp, caption={Example constraint description for an action constraint}, label={lst:constraint_description}]
Do not stack block1 on top of block2  
\end{lstlisting}
\begin{lstlisting}[language=lisp, caption={Annotated ground-truth Domain File for BlocksWorld-100 with action constraint description}, label={lst:blocksworld_df}]
(define (domain blocksworld)
;; CONSTRAINT Do not stack block1 on top of block2.
  (:requirements :strips)
(:predicates (clear ?x)
             (on-table ?x)
             (arm-empty)
             (holding ?x)
             (on ?x ?y)
;; BEGIN ADD
              (do-not-stack ?x ?y)
;; END ADD
             )

(:action pickup
  :parameters (?ob)
  :precondition (and (clear ?ob) (on-table ?ob) (arm-empty))
  :effect (and (holding ?ob) (not (clear ?ob)) (not (on-table ?ob)) 
               (not (arm-empty))))

(:action putdown
  :parameters  (?ob)
  :precondition (holding ?ob)
  :effect (and (clear ?ob) (arm-empty) (on-table ?ob) 
               (not (holding ?ob))))

(:action stack
  :parameters  (?ob ?underob)
  :precondition (and (clear ?underob) (holding ?ob) 
;; BEGIN ADD
                      (not (do-not-stack ?ob ?underob))
;; END ADD
                )
  :effect (and (arm-empty) (clear ?ob) (on ?ob ?underob)
               (not (clear ?underob)) (not (holding ?ob))))

(:action unstack
  :parameters  (?ob ?underob)
  :precondition (and (on ?ob ?underob) (clear ?ob) (arm-empty))
  :effect (and (holding ?ob) (clear ?underob)
               (not (on ?ob ?underob)) (not (clear ?ob)) (not (arm-empty)))))
\end{lstlisting}

\begin{lstlisting}[language=lisp, caption={Annotated ground-truth Problem File for BlocksWorld-100 with action constraint description}, label={lst:blocksworld-pf}]
(define (problem blocksworld-p50)
;; CONSTRAINT Do not stack block1 on top of block2.
  (:domain blocksworld)
  (:objects block1 block2 block3 block4 block5 block6 block7 )
  (:init 
    (on-table block3)
    (clear block3)
    (on-table block5)
    (clear block5)
    (on-table block7)
    (on block1 block7)
    (clear block1)
    (on-table block2)
    (clear block2)
    (on-table block6)
    (clear block6)
    (on-table block4)
    (clear block4)
    (arm-empty)
;; BEGIN ADD
    (do-not-stack block1 block2)
;; END ADD
  )
  (:goal (and 
    (on-table block5)
    (on block2 block5)
    (on-table block4)
    (on block7 block4)
    (on block6 block7)
    (on block1 block6)
    (on-table block3)
  ))
)
    \end{lstlisting}

\begin{lstlisting}[language=lisp, caption={Example Domain, Problem and Action Constraint Description for CoinCollector-100}, label={lst:coin-collector_description}]
;; Domain Description
You are standing in an environment with many rooms. Your task is to search the environment and find the coin.  Once you find the coin, take it.

The following actions are allowed: move from one room to another and take an item. 

;; Problem Description
You are in the kitchen.
To the east of the kitchen is the pantry.
To the west of the kitchen is the corridor.
There is a coin in the pantry.

;; Constraint Description
You must visit the corridor before the pantry.
\end{lstlisting}

\begin{lstlisting}[language=lisp, caption={Annotated ground-truth Domain File for CoinCollector problem with action constraint}, label={lst:coin-collector_df}]
(define (domain coin-collector)
;; CONSTRAINT You must visit the corridor before the pantry.
  (:requirements :strips :typing)
  (:types
    room
    direction
    item
  )
  (:predicates
    (at ?room - room)
    (connected ?room1 - room ?room2 - room ?direction - direction)
    (closed-door ?room1 - room ?room2 - room ?direction - direction)
    (location ?item - item ?room - room)
    (taken ?item - item)
    (is-reverse ?direction - direction ?reverse - direction)
;; BEGIN ADD
    (must-visit-before-next ?room - room)
    (next-room ?room - room)
    (visited-room)
;; END ADD
  )

  (:action move
    :parameters (?room1 - room ?room2 - room ?direction - direction)
    :precondition (and (at ?room1) (connected ?room1 ?room2 ?direction)
;; BEGIN ADD
    (not (and (next-room ?room2) (not (visited-room))))
;; END ADD
    )
    :effect (and (not (at ?room1)) (at ?room2)
;; BEGIN ADD
    (when (must-visit-before-next ?room2) (visited-room))
;; END ADD
    )
  )
  
  (:action take
    :parameters (?item - item ?room - room)
    :precondition (and (at ?room) (location ?item ?room) (not (taken ?item)))
    :effect (and (taken ?item) (not (location ?item ?room)))
  )
)
\end{lstlisting}

\begin{lstlisting}[language=lisp, caption={Annotated ground-truth Problem File for CoinCollector problem with action constraint}, label={lst: coin-collector_pf}]
(define (problem coin_collector_numLocations3_numDistractorItems0_seed26)
;; CONSTRAINT You must visit the corridor before the pantry.
  (:domain coin-collector)
  (:objects
    kitchen pantry corridor - room
    north south east west - direction
    coin - item
  )
  (:init
    (at kitchen)
    (connected kitchen pantry east)
    (connected kitchen corridor west)
    (connected pantry kitchen west)
    (connected corridor kitchen east)
    (location coin pantry)
    (is-reverse north south)
    (is-reverse south north)
    (is-reverse east west)
    (is-reverse west east)
;; BEGIN ADD
    (must-visit-before-next corridor)
    (next-room pantry)
;; END ADD
  )
  (:goal 
    (taken coin)
  )
)    
\end{lstlisting}

\begin{lstlisting}[language=lisp, caption={Annotated ground-truth Problem File for the XL BlocksWorld-100 problems with action constraint description}, label={lst:blocksworld-pf-xl}]
(define (problem blocksworld-p50)
;; CONSTRAINT Do not stack block1 on top of block2.
  (:domain blocksworld)
  (:objects block1 block2 block3 block4 block5 block6 block7 block8 block9 block10 block11 block12 block13 block14 block15 block16 block17 block18 block19 block20 block21 block22 block23 block24 block25 block26 block27 block28 block29 block30 block31 block32 block33 block34 block35 block36 block37 block38 block39 block40 block41 block42 block43 block44 block45 block46 block47 block48 block49 block50 )
  (:init 
    (on-table block10)
    (on block41 block10)
    (on block50 block41)
    (on block29 block50)
    (on block46 block29)
    (on block43 block46)
    (on block5 block43)
    (on block38 block5)
    (on block48 block38)
    (on block8 block48)
    (on block11 block8)
    (on block22 block11)
    (on block13 block22)
    (on block35 block13)
    (on block49 block35)
    (on block20 block49)
    (on block31 block20)
    (on block34 block31)
    (on block17 block34)
    (on block28 block17)
    (on block14 block28)
    (on block47 block14)
    (on block26 block47)
    (on block6 block26)
    (on block4 block6)
    (on block25 block4)
    (on block9 block25)
    (on block23 block9)
    (on block15 block23)
    (on block21 block15)
    (on block18 block21)
    (on block39 block18)
    (on block33 block39)
    (on block1 block33)
    (on block3 block1)
    (on block2 block3)
    (on block44 block2)
    (on block16 block44)
    (on block45 block16)
    (on block7 block45)
    (on block40 block7)
    (on block24 block40)
    (on block27 block24)
    (on block36 block27)
    (on block19 block36)
    (clear block19)
    (on-table block32)
    (clear block32)
    (on-table block42)
    (on block37 block42)
    (on block30 block37)
    (on block12 block30)
    (clear block12)
    (arm-empty)
;; BEGIN ADD
    (do-not-stack block1 block2)
;; END ADD
  )
  (:goal (and 
    (on-table block46)
    (on block45 block46)
    (on block37 block45)
    (on-table block30)
    (on block39 block30)
    (on-table block34)
    (on-table block49)
    (on-table block19)
    (on-table block38)
    (on-table block16)
    (on-table block25)
    (on-table block18)
    (on block5 block18)
    (on block17 block5)
    (on-table block42)
    (on-table block3)
    (on block32 block3)
    (on block35 block32)
    (on block26 block35)
    (on-table block12)
    (on-table block33)
    (on-table block14)
    (on-table block2)
    (on-table block23)
    (on block1 block23)
    (on block6 block1)
    (on-table block4)
    (on block20 block4)
    (on-table block48)
    (on-table block27)
    (on-table block31)
    (on-table block44)
    (on block47 block44)
    (on-table block43)
    (on-table block36)
    (on-table block7)
    (on-table block22)
    (on-table block28)
    (on-table block9)
    (on-table block11)
    (on-table block50)
    (on-table block24)
    (on-table block41)
    (on-table block8)
    (on-table block21)
    (on-table block40)
    (on-table block13)
    (on-table block15)
    (on-table block29)
    (on-table block10)
  ))
)
    \end{lstlisting}
\begin{lstlisting}[language=lisp, caption={Example constraint description for an action constraint for Mystery BlocksWorld}, label={lst:constraint_description-mbw}]
Do not perform action3 with object1 and object2.
\end{lstlisting}

\begin{lstlisting}[language=lisp, caption={Annotated ground-truth Domain File for Mystery BlocksWorld-100 with action constraint description}, label={lst:mystery_blocksworld-df}]
(define (domain mystery_blocksworld)
;; CONSTRAINT Do not perform action3 with object1 and object2.
  (:requirements :strips)
(:predicates (predicate1 ?x)
             (predicate2 ?x)
             (predicate3)
             (predicate4 ?x)
             (predicate5 ?x ?y)
;; BEGIN ADD
              (not-allowed ?x ?y)
;; END ADD
             )

(:action action1
  :parameters (?object1)
  :precondition (and (predicate1 ?object1) (predicate2 ?object1) (predicate3))
  :effect (and (predicate4 ?object1) (not (predicate1 ?object1)) (not (predicate2 ?object1)) 
               (not (predicate3))))

(:action action2
  :parameters  (?object1)
  :precondition (predicate4 ?object1)
  :effect (and (predicate1 ?object1) (predicate3) (predicate2 ?object1) 
               (not (predicate4 ?object1))))

(:action action3
  :parameters  (?object1 ?object2)
  :precondition (and (predicate1 ?object2) (predicate4 ?object1) 
;; BEGIN ADD
                      (not (not-allowed ?object1 ?object2))
;; END ADD
                )
  :effect (and (predicate3) (predicate1 ?object1) (predicate5 ?object1 ?object2)
               (not (predicate1 ?object2)) (not (predicate4 ?object1))))

(:action action4
  :parameters  (?object1 ?object2)
  :precondition (and (predicate5 ?object1 ?object2) (predicate1 ?object1) (predicate3))
  :effect (and (predicate4 ?object1) (predicate1 ?object2)
               (not (predicate5 ?object1 ?object2)) (not (predicate1 ?object1)) (not (predicate3)))))
    \end{lstlisting}

\begin{lstlisting}[language=lisp, caption={Annotated ground-truth Problem File for Mystery BlocksWorld-100 with action constraint description}, label={lst:mystery_blocksworld-pf}]
(define (problem mystery_blocksworld-p50)
;; CONSTRAINT Do not perform action3 with object1 and object2.
  (:domain mystery_blocksworld)
  (:objects object1 object2 object3 object4 object5 object6 object7 )
  (:init 
    (predicate2 object3)
    (predicate1 object3)
    (predicate2 object5)
    (predicate1 object5)
    (predicate2 object7)
    (predicate5 object1 object7)
    (predicate1 object1)
    (predicate2 object2)
    (predicate1 object2)
    (predicate2 object6)
    (predicate1 object6)
    (predicate2 object4)
    (predicate1 object4)
    (predicate3)
;; BEGIN ADD
    (not-allowed object1 object2)
;; END ADD
  )
  (:goal (and 
    (predicate2 object5)
    (predicate5 object2 object5)
    (predicate2 object4)
    (predicate5 object7 object4)
    (predicate5 object6 object7)
    (predicate5 object1 object6)
    (predicate2 object3)
  ))
)
    \end{lstlisting}

\begin{lstlisting}[language=lisp, caption={Prompt for LLM-as-Planner}, label={lst:prompt_LLM-as-Planner}]
You are a PDDL expert. Here is a game we are playing.
{domain_description}
{problem_description}
{constraint_description}

Write the plan that would solve this problem.

These are the available actions:
{available_actions}

Here is what the output should look like:
{example_answer}
    \end{lstlisting}

\begin{lstlisting}[language=lisp, caption={Prompt for LLM-as-PDDL-Formalizer, while generating the entire code}, label={lst:prompt_LLM-as-Formalizer-generate}]
You are a PDDL expert. Here is a game we are playing.
{domain_description}
{problem_description}
{constraint_description}

Write the domain and problem files in minimal PDDL.

These are the available actions:
{available_actions}
\end{lstlisting}

\begin{lstlisting}[language=lisp, caption={Prompt for LLM-as-PDDL-Formalizer, while first generating the non-constrained PDDL for the Edit setting}, label={lst:prompt_LLM-as-Formalizer-edit1}]
You are a PDDL expert. Here is a game we are playing.
{domain_description}
{problem_description}
Write the domain and problem files in minimal PDDL.

These are the available actions:
{available_actions}
\end{lstlisting}

\begin{lstlisting}[language=lisp, caption={Prompt for LLM-as-PDDL-Formalizer, while modifying the non-constrained PDDL for the Edit setting}, label={lst:prompt_LLM-as-Formalizer-edit2}]
You are a PDDL expert. Here is a PDDL domain and problem file.
{original_domain_file}
{original_problem_file}

Modify the PDDL files so that it satisfies the following constraint: {constraint_description}
These are the available actions:
{available_actions}

\end{lstlisting}

\begin{lstlisting}[language=lisp, caption={Prompt for LLM-as-PDDL-Formalizer, including revision}, label={lst:prompt_LLM-as-Formalizer-revision}]
You are a PDDL expert. The following domain and problem files have the error: {error}

{domain_file}
{problem_file}

Revise the PDDL to remove the error.
\end{lstlisting}

\begin{lstlisting}[language=lisp, caption={Prompt for LLM-as-PDDL3-Formalizer, while generating the entire code}, label={lst:prompt_pddl3-generate}]
You are a PDDL3 expert. Here is a game we are playing.
{domain_description}
{problem_description}
{constraint_description}

Write the domain and problem files in minimal PDDL3 code.

These are the available actions:
{available_actions}
\end{lstlisting}

\begin{lstlisting}[language=lisp, caption={Prompt for LLM-as-PDDL3-Formalizer, while first generating the non-constrained PDDL for the Edit setting}, label={lst:prompt_pddl3-edit1}]
You are a PDDL3 expert. Here is a game we are playing.
{domain_description}
{problem_description}
Write the domain and problem files in minimal PDDL3 code.

These are the available actions:
{available_actions}
\end{lstlisting}

\begin{lstlisting}[language=lisp, caption={Prompt for LLM-as-PDDL3-Formalizer, while modifying the non-constrained PDDL for the Edit setting}, label={lst:prompt_pddl3-edit2}]
You are a PDDL3 expert. Here is a PDDL3 domain and problem file.
{original_domain_file}
{original_problem_file}

Modify the files using PDDL3 so that it satisfies the following constraint: {constraint_description}
These are the available actions:
{available_actions}
\end{lstlisting}

\begin{lstlisting}[language=lisp, caption={Prompts for LLM-as-PDDL3-Formalizer, including revision}, label={lst:prompt_pddl3-revision}]
;; Error when running Compiler
You are a PDDL3 expert. The following PDDL3 were compiled and resulted in an error. Revise the PDDL3 to remove the error. Return a JSON object in the following format:
{{
\"domain file\": ...,
\"problem file\":...
}}

Domain File:
{original_domain_file}

Problem File:
{original_problem_file}

Error Message:
{compilation_error}

;; Error when running Solver
You are a PDDL3 expert. The following PDDL3 were compiled and ran through a solver, which resulted in an error. Revise the original PDDL3 to remove the error. Return a JSON object in the following format:
{{
\"domain file\": ...,
\"problem file\":...
}}

Original Domain File:
{original_domain_file}

Original Problem File:
{original_problem_file}

Compiled Domain File:
{compiled_domain_file}

Compiled Problem File:
{compiled_problem_file}

Error Message:
{solver_error}

\end{lstlisting}

\begin{lstlisting}[language=lisp, caption={Prompt for LLM-as-SMT-Formalizer (generate)}, label={lst:prompt_llm-as-smt-formalizer}]
You are a Z3 expert. Here is a domain and problem instance for {domain}.
{domain_description}
{problem_description}
{constraint_description}

Generate Python code that uses the Z3 Python API to solve this instance. These are the available actions:
{available_actions}

The output of your Python code should be a plan in the following format:{example_plan}

Set the number of steps allowed to 100.

\end{lstlisting}

\begin{lstlisting}[language=lisp, caption={Prompts for LLM-as-SMT-Formalizer (edit)}, label={lst:prompt_llm-as-smt-formalizer_edit}]
;; Step 1
You are a Z3 expert. Here is a domain and problem instance for {domain}.
{domain_description}
{problem_description}
Generate Python code that uses the Z3 Python API to solve this instance. These are the available actions:
{available_actions}

The output of your Python code should be a plan in the following format:
{example_plan}

Set the number of steps allowed to 100.

;; Step 2
You are a Z3 expert. Here is a {domain} instance written in Z3.
{original_python_code}

Modify the Z3 code so that it satisfies the following constraint: {constraint_description}
These are the available actions:
{available_actions}

The output of your Python code should be a plan in the following format:
{example_plan}
Set the number of steps allowed to 100.

\end{lstlisting}

\begin{lstlisting}[language=lisp, caption={Prompt for revision step for LLM-as-SMT-Formalizer}, label={lst:prompt_llm-as-smt-formalizer_revision}]
You are a Z3 expert. The following file have the error: {error}

{python_file}

Revise the Z3 code to remove the error.

\end{lstlisting}

\begin{lstlisting}[language=lisp, caption={Prompt for LLM-as-LTL-Formalizer}, label={lst:prompt_llm-as-ltl-formalizer}]
;; coin_collector_formula
  You are given a description of a grid world with rooms, adjacencies (north/south/east/west relations), 
  and possibly items (e.g. a coin) located in some room. 

  Your task is to convert this description into a set of LTL formulas for planning. 
  Decompose the problem into 5 parts: 
  1. Initial condition
  2. Occupancy (exactly one room at a time)
  3. Move dynamics (allowed moves)
  4. Pickup dynamics (pickup conditions)
  5. Goal (eventual condition)

  ### Rules:
  - Each room should be an atomic proposition in its name (e.g., `kitchen`, `bedroom`, etc.).
  - Exactly one room must be true at any time.
  - Legal moves: if you are in room A, the next state must be in one of the adjacent rooms or staying at the current room.
  - Legal pickups: if there is a coin in a room R, define an atomic proposition `has_coin`, and add these two rules to pickup dynamics for `has_coin`:
    - If the agent is in R and does not have the coin, then in the next state it may either gain the coin or stay without it.
    - If the agent is not in R and does not have the coin, then in the next state it must still not have the coin.
    - If the agent has the coin, it continues to have it in all future states.
    - if the coin is in room R, then picking it up must only change `has_coin` but not coincide with moving.
  - Initial condition: mark the starting room. `has_coin` is always false initially.
  - Goal: eventually `has_coin` must be true.

  ### Output format:
  Produce 4 formulas in Python strings:
  - `init = "..."` 
  - `occupancy = "..."` 
  - `move_dynamics = "..."` 
  - `pickup_dynamics = "..."` 
  - `goal = "..."` 

  ---
  [DESCRIPTION]
  ---
  Now generate the formulas:

;; coin_collector_adjacency
  You are given a description of a world with rooms and directional relations.

  Your task is to output the adjacency dictionary in Python format.

  ### Rules:
  - Use room names exactly as given in the text, lowercased and underscores instead of spaces.
  - Each key in the dictionary is a room.
  - Each value is another dictionary: {neighbor: direction}.
  - The direction must be from the perspective of the key room (north, south, east, west).
  - Always include both directions if they are implied. 
    Example: "To the north of the kitchen is the pantry" means:
        adjacency["kitchen"]["pantry"] = "north"
        adjacency["pantry"]["kitchen"] = "south"

  ### Output format:
  adjacency = {
  "room1": {"neighbor1": "direction", "neighbor2": "direction", ...},
  "room2": {...},
  ...
  }

  ---
  [DESCRIPTION]
  ---
  Now generate only the adjacency dictionary in the output format:

;; coin_collector_constraints
  You are given a set of atomic propositions and an environment description that represent a grid world planning problem.
  - Each room name is an atomic proposition, e.g.,[ROOMS]
  - The agent may pick up a coin, tracked by proposition "has_coin".
  Environment description:
  [PROBELM_DESCRIPTION]

  Your task: Given an additional natural language constraint, output the corresponding LTL formula that enforces it.

  ### Rules:
  - Always output formulas in the form of Python strings, e.g., "G(...)" or "F(...)".
  - Use the existing atomic propositions exactly as defined.
  - If the constraint mentions a room, use the room's name directly.
  - You should only output formulas for the new constraints.

  ### Examples:
  - If a constraint says "never enter ROOM", output `"G(!ROOM)"`.
  - If a constraint says "only move south", expand it into formulas of the form:
      For each room r, if its southern neighbor is s, then:
        G( r -> X(s) )
      and forbid all other neighbors in the X operator.
  - If a constraint says "only move south or west", expand similarly, listing all allowed successors explicitly.
  - If a constraint says "must visit ROOM_A before ROOM_B", encode it as `"(!ROOM_B) U ROOM_A"`.
  - If a constraint says "visit ROOM_A right after pick up coin", be careful about the number of X operators:
      If you are picking up the coin in the next timestep, visiting room a will happen in the following timestep: "G( (!has_coin & X(has_coin)) -> XX(ROOM_A) )"
  - Always output Python string formulas, one per constraint.
  - Do not introduce undefined macros.

  ### Formatting Examples:
  Natural language: "The agent must never enter the backyard."
  Output: "G(!backyard)"

  Natural language: "The agent must visit the bathroom before visiting the bedroom."
  Output: "(!bedroom) U bathroom"

  ---
  Constraint Description:
  [CONSTRAINTS_DESCRIPTION]
  Now, given these natural language constraints, output only the corresponding LTL formulas, without explanations:


\end{lstlisting}
\twocolumn

\begin{table*}[!t]
\centering
\small
\setlength{\tabcolsep}{3pt}
\renewcommand{\arraystretch}{1.1}
\resizebox{\textwidth}{!}{
\begin{tabular}{%
    p{4.2cm}  
    p{3.4cm}  
    p{5.0cm}  
    c         
    c         
}
\toprule
Formalizer Type & Method & Model & Correctness (Non-Constrained) & Correctness (Constrained) \\
\midrule

\multirow{7}{*}{LLM-as-Planner} & \multirow{7}{*}{N/A} & Gemini-3-Flash & 98 & 59 \\
& & DeepSeek-R1 & 91 & 54 \\
& & DeepSeek-V3 & 69 & 27 \\
& & DeepSeek-V3.2 (Thinking) & 92 & 48 \\
& & DeepSeek-V3.2 (Non-Thinking) & 47 & 8 \\
& & Qwen3-32B & 61 & 28 \\
& & Qwen2.5-32B & 3 & 2 \\
\bottomrule
\end{tabular}
}
\caption{Correctness (\%) with and without constraints on \bw using LLM-as-Planner.}
\label{tab: bw_LLM-as-Planner}
\end{table*}

\begin{table*}[!t]
\centering
\small
\setlength{\tabcolsep}{3pt}
\renewcommand{\arraystretch}{1.1}
\resizebox{\textwidth}{!}{
\begin{tabular}{%
    p{4.2cm}  
    p{3.4cm}  
    p{5.0cm}  
    c         
    c         
}
\toprule
Formalizer Type & Method & Model & Correctness (Non-Constrained) & Correctness (Constrained) \\
\midrule
\multirow{28}{*}{LLM-as-PDDL-Formalizer}
& \multirow{7}{*}{Generate} & Gemini-3-Flash & 97 & 60 \\
& & DeepSeek-R1 & 69 & 40 \\
& & DeepSeek-V3 & 39 & 11 \\
& & DeepSeek-V3.2 (Thinking) & 65 & 39 \\
& & DeepSeek-V3.2 (Non-Thinking) & 69 & 21 \\
& & Qwen3-32B & 16 & 16 \\
& & Qwen2.5-32B & 4 & 6 \\
\cmidrule(lr){2-5}
& \multirow{7}{*}{Generate + Revision} & Gemini-3-Flash & 97 & 60 \\
& & DeepSeek-R1 & 79 & 46 \\
& & DeepSeek-V3 & 71 & 23 \\
& & DeepSeek-V3.2 (Thinking) & 69 & 42 \\
& & DeepSeek-V3.2 (Non-Thinking) & 70 & 23 \\
& & Qwen3-32B & 29 & 20 \\
& & Qwen2.5-32B & 5 & 7 \\
\cmidrule(lr){2-5}
& \multirow{7}{*}{Edit} & Gemini-3-Flash & 97 & 74 \\
& & DeepSeek-R1 & 75 & 42 \\
& & DeepSeek-V3 & 41 & 22 \\
& & DeepSeek-V3.2 (Thinking) & 66 & 37 \\
& & DeepSeek-V3.2 (Non-Thinking) & 60 & 25 \\
& & Qwen3-32B & 27 & 14 \\
& & Qwen2.5-32B & 3 & 5 \\
\cmidrule(lr){2-5}
& \multirow{7}{*}{Edit + Revision} & Gemini-3-Flash & 97 & 76 \\
& & DeepSeek-R1 & 75 & 56 \\
& & DeepSeek-V3 & 41 & 30 \\
& & DeepSeek-V3.2 (Thinking) & 66 & 44 \\
& & DeepSeek-V3.2 (Non-Thinking) & 60 & 27 \\
& & Qwen3-32B & 27 & 24 \\
& & Qwen2.5-32B & 3 & 5 \\
\bottomrule
\end{tabular}
}
\caption{Correctness (\%) with and without constraints on \bw using LLM-as-PDDL-Formalizer.}
\label{tab: bw_pddl}
\end{table*}

\begin{table*}[!t]
\centering
\small
\setlength{\tabcolsep}{3pt}
\renewcommand{\arraystretch}{1.1}
\resizebox{\textwidth}{!}{
\begin{tabular}{%
    p{4.2cm}  
    p{3.4cm}  
    p{5.0cm}  
    c         
    c         
}
\toprule
Formalizer Type & Method & Model & Correctness (Non-Constrained) & Correctness (Constrained) \\
\midrule

\multirow{28}{*}{LLM-as-PDDL3-Formalizer}
& \multirow{7}{*}{Generate} & Gemini-3-Flash & 95 & 58 \\
& & DeepSeek-R1 & 61 & 29 \\
& & DeepSeek-V3 & 64 & 9 \\
& & DeepSeek-V3.2 (Thinking) & 71 & 39 \\
& & DeepSeek-V3.2 (Non-Thinking) & 49 & 9 \\
& & Qwen3-32B & 35 & 12 \\
& & Qwen2.5-32B & 5 & 2 \\
\cmidrule(lr){2-5}
& \multirow{7}{*}{Generate + Revision} & Gemini-3-Flash & 99 & 62 \\
& & DeepSeek-R1 & 63 & 33 \\
& & DeepSeek-V3 & 76 & 13 \\
& & DeepSeek-V3.2 (Thinking) & 76 & 45 \\
& & DeepSeek-V3.2 (Non-Thinking) & 85 & 11 \\
& & Qwen3-32B & 43 & 18 \\
& & Qwen2.5-32B & 12 & 5 \\
\cmidrule(lr){2-5}
& \multirow{7}{*}{Edit} & Gemini-3-Flash & 100 & 61 \\
& & DeepSeek-R1 & 61 & 32 \\
& & DeepSeek-V3 & 57 & 24 \\
& & DeepSeek-V3.2 (Thinking) & 73 & 34 \\
& & DeepSeek-V3.2 (Non-Thinking) & 49 & 9 \\
& & Qwen3-32B & 33 & 17 \\
& & Qwen2.5-32B & 5 & 6 \\
\cmidrule(lr){2-5}
& \multirow{7}{*}{Edit + Revision} & Gemini-3-Flash & 100 & 74 \\
& & DeepSeek-R1 & 61 & 39 \\
& & DeepSeek-V3 & 57 & 32 \\
& & DeepSeek-V3.2 (Thinking) & 73 & 43 \\
& & DeepSeek-V3.2 (Non-Thinking) & 49 & 24 \\
& & Qwen3-32B & 33 & 24 \\
& & Qwen2.5-32B & 5 & 7 \\

\bottomrule
\end{tabular}
}
\caption{Correctness (\%) with and without constraints on \bw using LLM-as-PDDL3-Formalizer.}
\label{tab: bw_pddl3}
\end{table*}

\begin{table*}[!t]
\centering
\small
\setlength{\tabcolsep}{3pt}
\renewcommand{\arraystretch}{1.1}
\resizebox{\textwidth}{!}{
\begin{tabular}{%
    p{4.2cm}  
    p{3.4cm}  
    p{5.0cm}  
    c         
    c         
}
\toprule
Formalizer Type & Method & Model & Correctness (Non-Constrained) & Correctness (Constrained) \\
\midrule
\multirow{28}{*}{LLM-as-SMT-Formalizer}
& \multirow{7}{*}{Generate} & Gemini-3-Flash & 42 & 24 \\
& & DeepSeek-R1 & 2 & 0 \\
& & DeepSeek-V3 & 1 & 2 \\
& & DeepSeek-V3.2 (Thinking) & 4 & 2 \\
& & DeepSeek-V3.2 (Non-Thinking) & 0 & 2 \\
& & Qwen3-32B & 0 & 1 \\
& & Qwen2.5-32B & 0 & 2 \\
\cmidrule(lr){2-5}
& \multirow{7}{*}{Generate + Revision} & Gemini-3-Flash & 53 & 31 \\
& & DeepSeek-R1 & 28 & 13 \\
& & DeepSeek-V3 & 1 & 6 \\
& & DeepSeek-V3.2 (Thinking) & 36 & 18 \\
& & DeepSeek-V3.2 (Non-Thinking) & 0 & 8 \\
& & Qwen3-32B & 2 & 13 \\
& & Qwen2.5-32B & 0 & 4 \\
\cmidrule(lr){2-5}
& \multirow{7}{*}{Edit} & Gemini-3-Flash & 43 & 22 \\
& & DeepSeek-R1 & 1 & 1 \\
& & DeepSeek-V3 & 0 & 3 \\
& & DeepSeek-V3.2 (Thinking) & 6 & 6 \\
& & DeepSeek-V3.2 (Non-Thinking) & 0 & 3 \\
& & Qwen3-32B & 1 & 2 \\
& & Qwen2.5-32B & 1 & 2 \\
\cmidrule(lr){2-5}
& \multirow{7}{*}{Edit + Revision} & Gemini-3-Flash & 43 & 26 \\
& & DeepSeek-R1 & 1 & 22 \\
& & DeepSeek-V3 & 0 & 3 \\
& & DeepSeek-V3.2 (Thinking) & 6 & 31 \\
& & DeepSeek-V3.2 (Non-Thinking) & 0 & 13 \\
& & Qwen3-32B & 1 & 13 \\
& & Qwen2.5-32B & 1 & 5 \\
\bottomrule
\end{tabular}
}
\caption{Correctness (\%) with and without constraints on \bw using LLM-as-SMT-Formalizer.}
\label{tab: bw_smt}
\end{table*}

\begin{table*}[!t]
\centering
\small
\setlength{\tabcolsep}{3pt}
\renewcommand{\arraystretch}{1.1}
\resizebox{\textwidth}{!}{
\begin{tabular}{%
    p{4.2cm}  
    p{3.4cm}  
    p{5.0cm}  
    c         
    c         
}
\toprule
Formalizer Type & Method & Model & Correctness (Non-Constrained) & Correctness (Constrained) \\
\midrule

\multirow{7}{*}{LLM-as-Planner} & \multirow{7}{*}{N/A} & Gemini-3-Flash & 93 & 57 \\
& & DeepSeek-R1 & 89 & 49 \\
& & DeepSeek-V3 & 80 & 38 \\
& & DeepSeek-V3.2 (Thinking) & 84 & 48 \\
& & DeepSeek-V3.2 (Non-Thinking) & 76 & 38 \\
& & Qwen3-32B & 96 & 52 \\
& & Qwen2.5-32B & 79 & 36 \\
\bottomrule
\end{tabular}
}
\caption{Correctness (\%) with and without constraints on \coin using LLM-as-Planner.}
\label{tab: cc_planner}
\end{table*}

\begin{table*}[!t]
\centering
\small
\setlength{\tabcolsep}{3pt}
\renewcommand{\arraystretch}{1.1}
\resizebox{\textwidth}{!}{
\begin{tabular}{%
    p{4.2cm}  
    p{3.4cm}  
    p{5.0cm}  
    c         
    c         
}
\toprule
Formalizer Type & Method & Model & Correctness (Non-Constrained) & Correctness (Constrained) \\
\midrule
\multirow{28}{*}{LLM-as-PDDL-Formalizer}
& \multirow{7}{*}{Generate} & Gemini-3-Flash & 90 & 52 \\
& & DeepSeek-R1 & 65 & 26 \\
& & DeepSeek-V3 & 64 & 24 \\
& & DeepSeek-V3.2 (Thinking) & 52 & 23 \\
& & DeepSeek-V3.2 (Non-Thinking) & 68 & 30 \\
& & Qwen3-32B & 45 & 20 \\
& & Qwen2.5-32B & 3 & 3 \\
\cmidrule(lr){2-5}
& \multirow{7}{*}{Generate + Revision} & Gemini-3-Flash & 91 & 53 \\
& & DeepSeek-R1 & 75 & 34 \\
& & DeepSeek-V3 & 65 & 25 \\
& & DeepSeek-V3.2 (Thinking) & 52 & 24 \\
& & DeepSeek-V3.2 (Non-Thinking) & 68 & 33 \\
& & Qwen3-32B & 48 & 21 \\
& & Qwen2.5-32B & 5 & 4 \\
\cmidrule(lr){2-5}
& \multirow{7}{*}{Edit} & Gemini-3-Flash & 88 & 57 \\
& & DeepSeek-R1 & 62 & 39 \\
& & DeepSeek-V3 & 64 & 21 \\
& & DeepSeek-V3.2 (Thinking) & 58 & 38 \\
& & DeepSeek-V3.2 (Non-Thinking) & 76 & 26 \\
& & Qwen3-32B & 50 & 18 \\
& & Qwen2.5-32B & 4 & 2 \\
\cmidrule(lr){2-5}
& \multirow{7}{*}{Edit + Revision} & Gemini-3-Flash & 88 & 57 \\
& & DeepSeek-R1 & 62 & 41 \\
& & DeepSeek-V3 & 64 & 21 \\
& & DeepSeek-V3.2 (Thinking) & 58 & 39 \\
& & DeepSeek-V3.2 (Non-Thinking) & 76 & 29 \\
& & Qwen3-32B & 50 & 24 \\
& & Qwen2.5-32B & 4 & 2 \\
\bottomrule
\end{tabular}
}
\caption{Correctness (\%) with and without constraints on \coin using LLM-as-PDDL-Formalizer.}
\label{tab: cc_pddl}
\end{table*}

\begin{table*}[!t]
\centering
\small
\setlength{\tabcolsep}{3pt}
\renewcommand{\arraystretch}{1.1}
\resizebox{\textwidth}{!}{
\begin{tabular}{%
    p{4.2cm}  
    p{3.4cm}  
    p{5.0cm}  
    c         
    c         
}
\toprule
Formalizer Type & Method & Model & Correctness (Non-Constrained) & Correctness (Constrained) \\
\midrule

\multirow{28}{*}{LLM-as-PDDL3-Formalizer}
& \multirow{7}{*}{Generate} & Gemini-3-Flash & 93 & 48 \\
& & DeepSeek-R1 & 54 & 21 \\
& & DeepSeek-V3 & 13 & 8 \\
& & DeepSeek-V3.2 (Thinking) & 50 & 24 \\
& & DeepSeek-V3.2 (Non-Thinking) & 24 & 8 \\
& & Qwen3-32B & 24 & 8 \\
& & Qwen2.5-32B & 2 & 0 \\
\cmidrule(lr){2-5}
& \multirow{7}{*}{Generate + Revision} & Gemini-3-Flash & 99 & 62 \\
& & DeepSeek-R1 & 55 & 30 \\
& & DeepSeek-V3 & 49 & 11 \\
& & DeepSeek-V3.2 (Thinking) & 54 & 30 \\
& & DeepSeek-V3.2 (Non-Thinking) & 46 & 13 \\
& & Qwen3-32B & 58 & 26 \\
& & Qwen2.5-32B & 21 & 8 \\
\cmidrule(lr){2-5}
& \multirow{7}{*}{Edit} & Gemini-3-Flash & 92 & 45 \\
& & DeepSeek-R1 & 60 & 30 \\
& & DeepSeek-V3 & 22 & 8 \\
& & DeepSeek-V3.2 (Thinking) & 63 & 34 \\
& & DeepSeek-V3.2 (Non-Thinking) & 23 & 9 \\
& & Qwen3-32B & 21 & 7 \\
& & Qwen2.5-32B & 0 & 0 \\
\cmidrule(lr){2-5}
& \multirow{7}{*}{Edit + Revision} & Gemini-3-Flash & 92 & 59 \\
& & DeepSeek-R1 & 60 & 32 \\
& & DeepSeek-V3 & 22 & 17 \\
& & DeepSeek-V3.2 (Thinking) & 63 & 39 \\
& & DeepSeek-V3.2 (Non-Thinking) & 23 & 20 \\
& & Qwen3-32B & 21 & 25 \\
& & Qwen2.5-32B & 0 & 8 \\

\bottomrule
\end{tabular}
}
\caption{Correctness (\%) with and without constraints on \coin using LLM-as-PDDL3-Formalizer.}
\label{tab: cc_pddl3}
\end{table*}

\begin{table*}[!t]
\centering
\small
\setlength{\tabcolsep}{3pt}
\renewcommand{\arraystretch}{1.1}
\resizebox{\textwidth}{!}{
\begin{tabular}{%
    p{4.2cm}  
    p{3.4cm}  
    p{5.0cm}  
    c         
    c         
}
\toprule
Formalizer Type & Method & Model & Correctness (Non-Constrained) & Correctness (Constrained) \\
\midrule
\multirow{28}{*}{LLM-as-SMT-Formalizer}
& \multirow{7}{*}{Generate} & Gemini-3-Flash & 75 & 40 \\
& & DeepSeek-R1 & 45 & 23 \\
& & DeepSeek-V3 & 42 & 23 \\
& & DeepSeek-V3.2 (Thinking) & 36 & 24 \\
& & DeepSeek-V3.2 (Non-Thinking) & 30 & 14 \\
& & Qwen3-32B & 29 & 18 \\
& & Qwen2.5-32B & 1 & 7 \\
\cmidrule(lr){2-5}
& \multirow{7}{*}{Generate + Revision} & Gemini-3-Flash & 82 & 48 \\
& & DeepSeek-R1 & 77 & 39 \\
& & DeepSeek-V3 & 58 & 32 \\
& & DeepSeek-V3.2 (Thinking) & 46 & 30 \\
& & DeepSeek-V3.2 (Non-Thinking) & 38 & 21 \\
& & Qwen3-32B & 49 & 32 \\
& & Qwen2.5-32B & 1 & 11 \\
\cmidrule(lr){2-5}
& \multirow{7}{*}{Edit} & Gemini-3-Flash & 64 & 41 \\
& & DeepSeek-R1 & 43 & 29 \\
& & DeepSeek-V3 & 39 & 24 \\
& & DeepSeek-V3.2 (Thinking) & 40 & 24 \\
& & DeepSeek-V3.2 (Non-Thinking) & 28 & 14 \\
& & Qwen3-32B & 28 & 17 \\
& & Qwen2.5-32B & 1 & 3 \\
\cmidrule(lr){2-5}
& \multirow{7}{*}{Edit + Revision} & Gemini-3-Flash & 64 & 46 \\
& & DeepSeek-R1 & 43 & 47 \\
& & DeepSeek-V3 & 39 & 33 \\
& & DeepSeek-V3.2 (Thinking) & 40 & 29 \\
& & DeepSeek-V3.2 (Non-Thinking) & 28 & 17 \\
& & Qwen3-32B & 28 & 24 \\
& & Qwen2.5-32B & 1 & 7 \\
\bottomrule
\end{tabular}
}
\caption{Correctness (\%) with and without constraints on \coin using LLM-as-SMT-Formalizer.}
\label{tab: cc_smt}
\end{table*}

\begin{table*}[!t]
\centering
\small
\setlength{\tabcolsep}{3pt}
\renewcommand{\arraystretch}{1.1}
\resizebox{\textwidth}{!}{
\begin{tabular}{%
    p{4.2cm}  
    p{3.4cm}  
    p{5.0cm}  
    c         
    c         
}
\toprule
Formalizer Type & Method & Model & Correctness (Non-Constrained) & Correctness (Constrained) \\
\midrule

\multirow{7}{*}{LLM-as-LTL-Formalizer}
& \multirow{7}{*}{Generate} & Gemini-3-Flash & 97 & 36 \\
& & DeepSeek-R1 & 100 & 44 \\
& & DeepSeek-V3 & 74 & 33 \\
& & DeepSeek-V3.2 (Thinking) & 95 & 35 \\
& & DeepSeek-V3.2 (Non-Thinking) & 82 & 26 \\
& & Qwen3-32B & 0 & 7 \\
& & Qwen2.5-32B & 18 & 11 \\

\bottomrule
\end{tabular}
}
\caption{Correctness (\%) with and without constraints on \coin using LLM-as-LTL-Formalizer.}
\label{tab: cc_ltl}
\end{table*}

\begin{table*}[!t]
\centering
\small
\setlength{\tabcolsep}{3pt}
\renewcommand{\arraystretch}{1.1}
\resizebox{\textwidth}{!}{
\begin{tabular}{%
    p{4.2cm}  
    p{3.4cm}  
    p{5.0cm}  
    c         
    c         
}
\toprule
Formalizer Type & Method & Model & Correctness (Non-Constrained) & Correctness (Constrained) \\
\midrule

\multirow{7}{*}{LLM-as-Planner} & \multirow{7}{*}{N/A} & Gemini-3-Flash & 97 & 42 \\
& & DeepSeek-R1 & 44 & 31 \\
& & DeepSeek-V3 & 10 & 8 \\
& & DeepSeek-V3.2 (Thinking) & 68 & 25 \\
& & DeepSeek-V3.2 (Non-Thinking) & 4 & 2 \\
& & Qwen3-32B & 4 & 3 \\
& & Qwen2.5-32B & 1 & 1 \\
\midrule

\multirow{14}{*}{LLM-as-PDDL-Formalizer}
& \multirow{7}{*}{Generate} & Gemini-3-Flash & 72 & 29 \\
& & DeepSeek-R1 & 48 & 13 \\
& & DeepSeek-V3 & 22 & 5 \\
& & DeepSeek-V3.2 (Thinking) & 29 & 18 \\
& & DeepSeek-V3.2 (Non-Thinking) & 48 & 11 \\
& & Qwen3-32B & 11 & 12 \\
& & Qwen2.5-32B & 0 & 4 \\
\cmidrule(lr){2-5}
& \multirow{7}{*}{Edit} & Gemini-3-Flash & 71 & 43 \\
& & DeepSeek-R1 & 41 & 19 \\
& & DeepSeek-V3 & 28 & 12 \\
& & DeepSeek-V3.2 (Thinking) & 26 & 24 \\
& & DeepSeek-V3.2 (Non-Thinking) & 49 & 27 \\
& & Qwen3-32B & 7 & 5 \\
& & Qwen2.5-32B & 1 & 4 \\

\bottomrule
\end{tabular}
}
\caption{Correctness (\%) with and without constraints on \bwxl using LLM-as-Planner and LLM-as-PDDL-Formalizer.}
\label{tab: all blocksworld-xl}
\end{table*}

\begin{table*}[!ht]
\centering
\small
\setlength{\tabcolsep}{3pt}
\renewcommand{\arraystretch}{1.1}
\resizebox{\textwidth}{!}{
\begin{tabular}{%
    p{4.2cm}  
    p{3.4cm}  
    p{5.0cm}  
    c         
    c         
}
\toprule
Formalizer Type & Method & Model & Correctness (Non-Constrained) & Correctness (Constrained) \\
\midrule

\multirow{7}{*}{LLM-as-Planner} & \multirow{7}{*}{N/A} & Gemini-3-Flash & 88 & 25 \\
& & DeepSeek-R1 & 24 & 15 \\
& & DeepSeek-V3 & 11 & 15 \\
& & DeepSeek-V3.2 (Thinking) & 33 & 12 \\
& & DeepSeek-V3.2 (Non-Thinking) & 1 & 1 \\
& & Qwen3-32B & 10 & 4 \\
& & Qwen2.5-32B & 1 & 1 \\
\midrule

\multirow{16}{*}{LLM-as-PDDL-Formalizer}
& \multirow{4}{*}{Generate} & Gemini-3-Flash & 100 & 31 \\
& & DeepSeek-R1 & 78 & 20 \\
& & DeepSeek-V3 & 63 & 15 \\
& & DeepSeek-V3.2 (Thinking) & 100 & 30 \\
& & DeepSeek-V3.2 (Non-Thinking) & 99 & 19 \\
& & Qwen3-32B & 85 & 19 \\
& & Qwen2.5-32B & 89 & 19 \\
\cmidrule(lr){2-5}
& \multirow{7}{*}{Edit} & Gemini-3-Flash & 100 & 55 \\
& & DeepSeek-R1 & 82 & 23 \\
& & DeepSeek-V3 & 66 & 15 \\
& & DeepSeek-V3.2 (Thinking) & 99 & 33 \\
& & DeepSeek-V3.2 (Non-Thinking) & 99 & 25 \\
& & Qwen3-32B & 84 & 27 \\
& & Qwen2.5-32B & 95 & 18 \\

\bottomrule
\end{tabular}
}
\caption{Correctness (\%) with and without constraints on \mbw using LLM-as-Planner and LLM-as-PDDL-Formalizer.}
\label{tab: all mbw}
\end{table*}

\end{document}